\pdfoutput=1
\PassOptionsToPackage{table}{xcolor}

\documentclass[11pt]{article}

\usepackage[final]{acl}
\usepackage{booktabs}
\usepackage{times}
\usepackage{latexsym}
\usepackage{float}   
\usepackage{stfloats}   
\usepackage{multirow}
\usepackage{comment}
\usepackage{amsfonts}

\definecolor{softred}{RGB}{220, 50, 47}
\definecolor{softblue}{RGB}{38, 139, 210}

\usepackage[T1]{fontenc}

\usepackage[utf8]{inputenc}

\usepackage{microtype}

\usepackage{inconsolata}

\usepackage{graphicx}

\usepackage{amsmath}
\usepackage{xspace}

\def\secref#1{\S\ref{sec:#1}}
\def\seclabel#1{\label{sec:#1}}

\title{Your Pretrained Model Tells the Difficulty Itself: A Self-Adaptive Curriculum Learning Paradigm for Natural Language Understanding}

\author{
Qi Feng\textsuperscript{1}\thanks{Equal contribution.} \quad
Yihong Liu\textsuperscript{1,2}\footnotemark[1] \quad
Hinrich Schütze\textsuperscript{1,2} \\
\textsuperscript{1}Center for Information and Language Processing, LMU Munich \\
\textsuperscript{2}Munich Center for Machine Learning (MCML) \\
\texttt{fengqi928@outlook.com}
}

\begin{document}
\maketitle
\begin{abstract}

Curriculum learning is a widely adopted training strategy in natural language processing (NLP), where models are exposed to examples organized by increasing \emph{difficulty} to enhance learning efficiency and performance. 
However, most existing approaches rely on manually defined difficulty metrics -- such as text length -- which may not accurately reflect the model’s own perspective. 
To overcome this limitation, we present a self-adaptive curriculum learning paradigm that prioritizes fine-tuning examples based on difficulty scores predicted by pre-trained language models (PLMs) themselves. 
Building on these scores, we explore various training strategies that differ in the ordering of examples for the fine-tuning: from easy-to-hard, hard-to-easy, to mixed sampling. 
We evaluate our method on four natural language understanding (NLU) datasets covering both binary and multi-class classification tasks.
Experimental results show that our approach leads to faster convergence and improved performance compared to standard random sampling. 
We make our code publicly available.\footnote{\url{https://github.com/alitanokiki/self-adaptive-curriculum-nlu-acl2025}} 

\end{abstract}

\section{Introduction}

Although large language models (LLMs) are highly valued in the NLP community for their broad capabilities \citep{naveed2024llm,Chang2024llm-eval}, their substantial computational cost often makes them impractical for many real-world scenarios -- particularly for simple classification tasks that require rapid responses or deployment on resource-constrained infrastructure \citep{bai2024efficiency,Cunningham2024Efficient}. 
As a result, \emph{task-specific} NLP models -- those pre-trained and subsequently fine-tuned on labeled data for specific tasks, e.g., sentiment analysis -- remain highly relevant \citep{zhao-etal-2024-syntheval}. 
While many studies have focused on enhancing the effectiveness of pre-training \citep{du-etal-2021-self,yu2022jaket,liu2024langsamp,hu2024Knowledge}, the high resource demands of this stage make it more practical to instead develop improved fine-tuning strategies \citep{xu-etal-2020-curriculum,chen-etal-2021-hiddencut,Hu2022lora,Ding2023parameter}.

\begin{figure}[t]
    \centering
    \setlength{\belowcaptionskip}{-0.5cm}
    \includegraphics[width=0.45\textwidth]{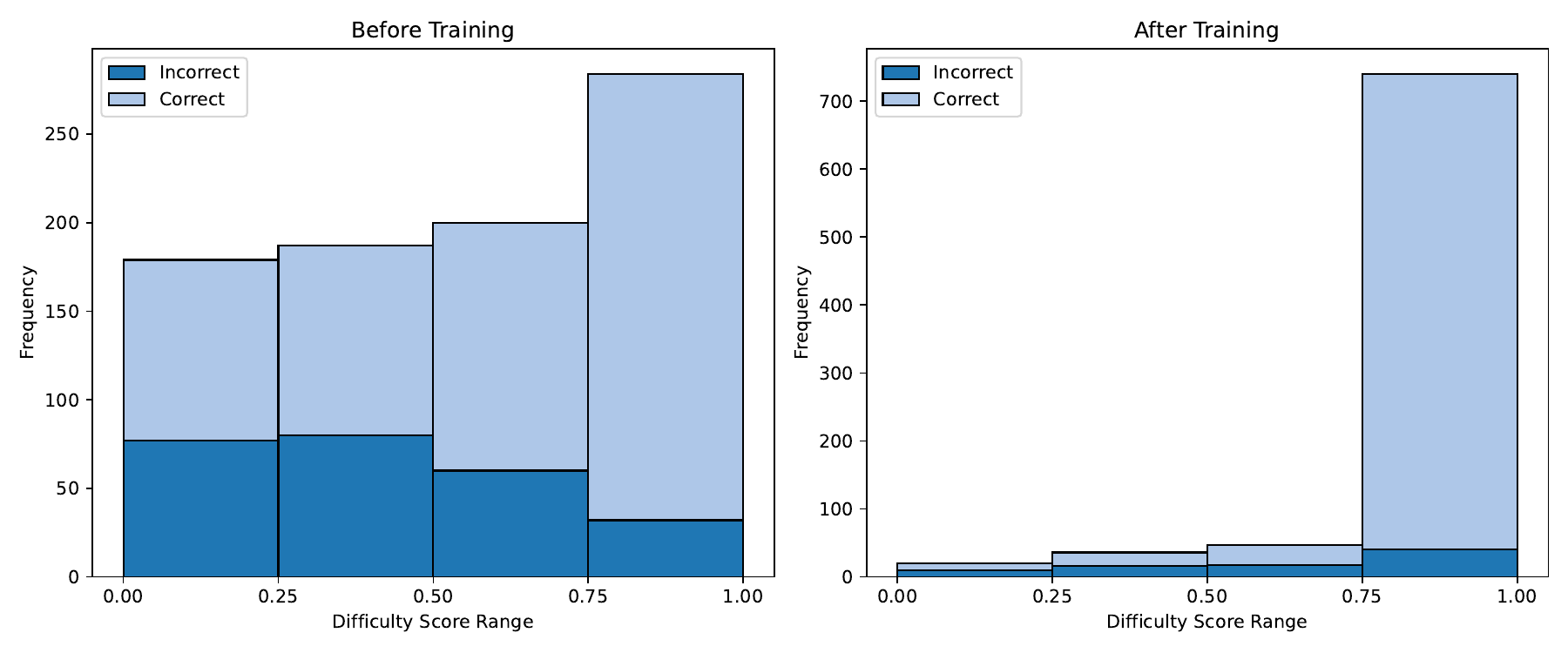}
    \caption{Frequencies of samples being incorrectly (\textcolor{blue!70!black}{dark blue}) and correctly classified (\textcolor{blue!30}{light blue}) by BERT before and after 1 epoch of training. The model tends to make worse decisions when samples are difficult and better decisions when they are easy. Note that a sample with a difficulty score of 0 is the most difficult one.}
    \label{fig:scores}
\end{figure}

One important class of fine-tuning strategies centers around the concept of \emph{curriculum} -- a process inspired by human learning. 
\emph{Curriculum Learning}, first introduced by \citet{bengio2009curriculum} in the general machine learning domain, has since demonstrated effectiveness in NLP tasks as well \citep{xu-etal-2020-curriculum,zhu-etal-2021-combining-curriculum-learning,maharana-bansal-2022-curriculum,ranaldi-etal-2023-modeling,Gao2024Confucius}. 
This paradigm involves structuring training data from simpler to more complex examples, enabling models to build knowledge incrementally and learn more efficiently. 
A central challenge in applying curriculum learning lies in defining \emph{difficulty}. 
Most prior work estimates difficulty using surface-level features such as sentence length or word rarity \citep{platanios-etal-2019-competence,xu-etal-2020-curriculum,ranaldi-etal-2023-modeling}. 
However, these metrics may not align with the model’s internal understanding -- especially for PLMs capable of capturing deeper semantic attributes like irony or ambiguity thanks to massive pre-training. 
Moreover, the assumption that training should always progress from easy to hard is debatable; 
models may benefit from early exposure to difficult examples or from revisiting easier ones in training to mitigate forgetting \citep{Kirkpatrick2017catastrophic,Ke2021Forgetting,huang-etal-2024-mitigating}.

To this end, we propose a self-adaptive curriculum learning paradigm that explores various sampling strategies driven by the model’s own confidence. 
Rather than relying on manually defined difficulty heuristics based on the surface feature of an example, we leverage the PLM itself to compute a difficulty score -- specifically, a confidence measure that reflects how certain the model is when classifying an example using a prompt template and a verbalizer component \citep{schick-schutze-2021-exploiting}. 
For each example, we define its difficulty as the maximum absolute difference among the predicted class probabilities, where a smaller difference indicates greater uncertainty (i.e., higher difficulty).
Since this computation requires no parameter updates, it can be performed efficiently across the dataset.
Once difficulty scores are computed, we sort the examples in ascending or descending order and explore three categories of sampling strategies:
\textbf{\emph{Naive sequential sampling}}: examples are selected in order from easiest to hardest, or in reverse.
\textbf{\emph{Probability-based sampling}}: examples are sampled probabilistically, with sampling probabilities defined based on their difficulty ranks.
\textbf{\emph{Partitioned batch sampling}}: examples are divided into easy and hard groups, and batches are formed by sampling from both partitions during fine-tuning.

To validate our proposed methodology, we conduct extensive experiments on four NLU datasets covering both binary and multi-class classification tasks, including sentiment analysis, hate speech detection, and natural language inference. 
We show that the difficulty scores predicted by the PLM itself serve as a reliable proxy for model uncertainty -- examples with lower scores are much more likely to be misclassified, as shown in Figure~\ref{fig:scores}.
Moreover, our sampling strategies yield competitive or superior performance compared to standard random sampling in the full-dataset fine-tuning setting.
In the few-shot fine-tuning setting, our methods generally outperform the baseline methods, demonstrating strong generalization and robustness.
Our contributions are as follows:

(i) We propose a self-adaptive curriculum paradigm that prioritizes fine-tuning examples based on difficulty scores predicted by the PLM itself.
(ii) We propose three categories of sampling strategies based on ranked lists of examples according to their difficulty scores.
(iii) We empirically validate our approach on four diverse NLU tasks, achieving strong results in both full-dataset and few-shot fine-tuning scenarios.

\section{Related Work}

\subsection{Sampling Strategies}
Traditional random sampling methods, though widely used, often fail to make the model learning more effective.
Therefore, more advanced sampling strategies have been explored, including strategies with stratified sampling \citep{b5930ac0-b2f3-3b38-aec0-0a4828ad91f1,qian-etal-2009-semi}, multistage sampling \citep{nadeem-etal-2020-systematic}, adaptive ranking-based sampling \citep{song-etal-2022-adaptive} and class balancing techniques such as balanced data sampling \citep{shao-etal-2024-balanced}. 
Active learning (AL) selects the most informative instances for annotation \cite{DBLP:journals/corr/LewisG94} to better leverage unlabeled data, with recent strategies including uncertainty-based sampling \citep{yu2022actune}, cold-start AL via masked language modeling loss \citep{yuan-etal-2020-cold}, self-active learning for multilingual settings \citep{dossou-etal-2022-afrolm}, and hybrid AL combining uncertainty and diversity \citep{azeemi-etal-2025-label}. 
A comprehensive survey of AL in NLP is provided by \citet{zhang-etal-2022-survey}. Adaptive sampling techniques, which dynamically adjust sample selection during training, recent research includes difficulty-aware negative sampling \citep{li-etal-2019-sampling}, hard negative mining in extreme classification \citep{dahiya2023ngame}, and class-adaptive re-sampling to mitigate false negatives in weak supervision \citep{tan-etal-2023-class}.

\subsection{Curriculum Learning}

Curriculum learning (CL) \citep{bengio2009curriculum} defines the difficulty of the sample and improves model convergence and performance by ordering training samples from easy to hard \citep{soviany2022curriculumlearningsurvey}. In NLP, it can be implemented by sorting and sampling sentences based on features such as sentence length or word rarity \citep{platanios-etal-2019-competence}. However, empirical results suggest that
such heuristics may offer limited benefits over
random sampling \citep{surkov-etal-2022-data}.
Beyond manual annotations or simple heuristics, CL variants differ in how they define difficulty and structure training. Teacher-student CL ranks samples via an external model \citep{xu-etal-2020-curriculum, soviany2022curriculumlearningsurvey}, while self-paced CL allows models to select samples based on their internal progress \citep{jiang2015self}. 
Competence-based CL introduces a formal notion of model competence, and dynamically filters training samples \citep{platanios-etal-2019-competence}. \citet{wu2021curriculawork} examine whether curriculum or anti-curriculum ordering improves training, and find limited benefits over random sampling in standard settings. 
Beyond these mainstream variants, more recent work has extended curriculum learning into various specialized settings, including combining CL with active learning \citep{jafarpour-etal-2021-active}, dual CL, which handles positive and negative samples separately \citep{Zhu_2022}, and curriculum contrastive learning for knowledge graph entity typing \citep{wang-etal-2025-knowledge}. Recent work also applies curriculum learning to code language models by defining difficulty through static complexity measures \citep{nair-etal-2024-curriculum}.
Some methods follow curriculum principles without being explicitly framed as curriculum learning \citep{mindermann2022prioritizedtrainingpointslearnable, thakkar-etal-2023-self}. 
In contrast to this line of work, we propose a CL framework relying on the difficulty predicted by the model itself, without relying on external models, metrics, or annotations.

\subsection{Prompt-Based Fine-Tuning}

Prompt-based Fine-tuning (PFT) has emerged as a powerful approach for adapting PLMs to downstream tasks, particularly in zero-shot and few-shot scenarios \citep{schick-schutze-2021-exploiting, schick-schutze-2021-just, schick-schutze-2021-shot, le-scao-rush-2021-many, gao-etal-2021-making, jin-etal-2022-good, 10.1145/3603168, ma-etal-2023-prompt, ullah-etal-2023-comparing, xie-li-2024-discriminative}
.
An important early stage of PFT research was marked by Pattern-Exploiting Training (PET), proposed by \citet{schick-schutze-2021-just}.
Building on this, \citet{schick-schutze-2021-exploiting, schick-schutze-2021-shot} further explored key factors such as prompt design, verbalizer selection, and self-training strategies, and extended PET to text generation tasks. 
In PFT, verbalizers can either be manually crafted or automatically optimized \citep{shin-etal-2020-autoprompt,schick-schutze-2021-exploiting}. 
Recent work has further extended PFT beyond monolingual settings to multilingual and cross-lingual tasks \citep{hu-etal-2022-knowledgeable, Ye_2022, wang-etal-2022-automatic, ma-etal-2023-prompt}.
 
While early studies primarily focused on single-label classification, more recent efforts have adapted PFT to more complex settings such as multi-label classification \citep{yang-etal-2022-knowledge-injected}. Recent work has also addressed semantic inconsistency and representation degeneration in prompt-based fine-tuning, proposing methods such as semantic consistency modeling \citep{xie-li-2024-discriminative} and contrastive learning frameworks \citep{zhao-etal-2024-representation}.

\begin{figure}[t]
    \centering
    \setlength{\abovecaptionskip}{-0.01cm}
    \includegraphics[width=0.99\linewidth]{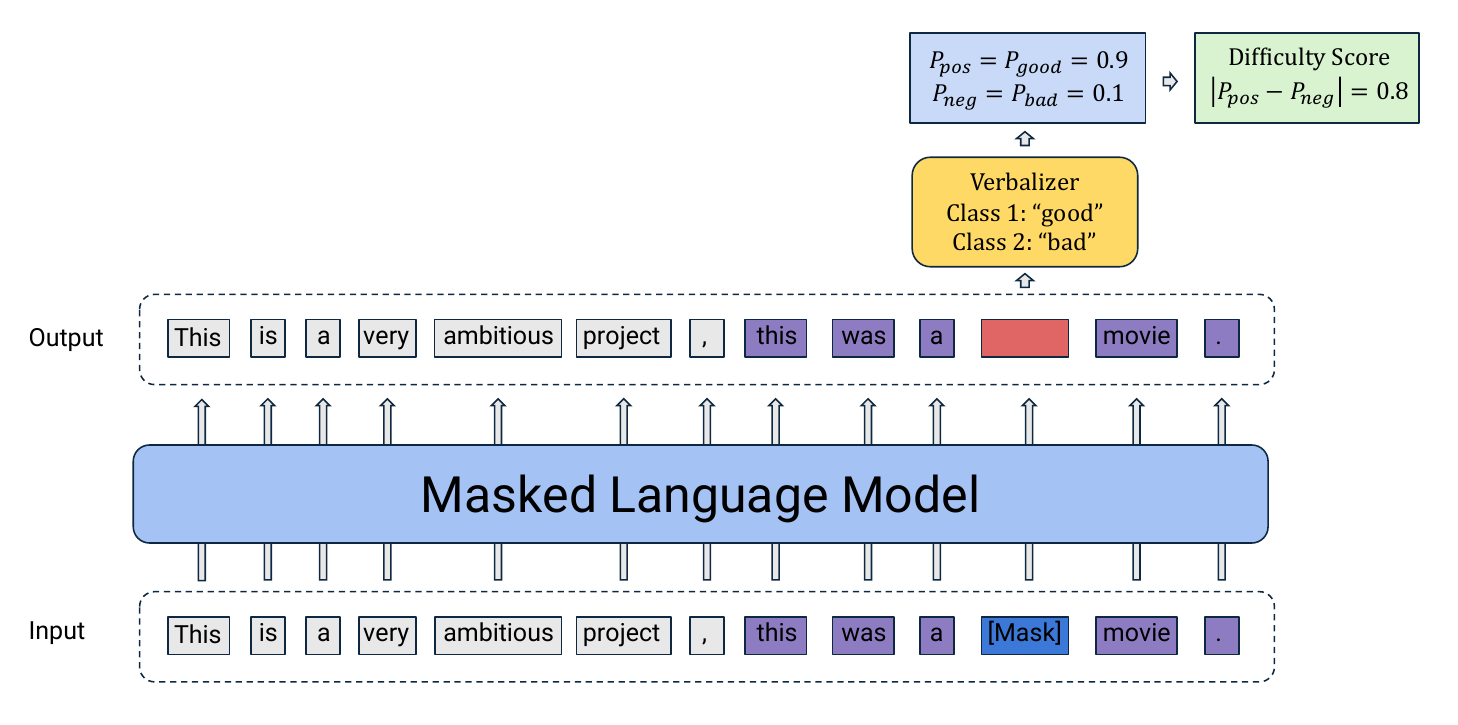}
    \caption{Illustration of the proposed difficulty scoring approach using masked language modeling and a verbalizer. The input sentence is processed to predict the masked token, and the resulting token probabilities are mapped to class labels through a verbalizer. In this example, the tokens ``good'' and ``bad'' represent the positive and negative classes, respectively.
    The difficulty score is then computed as the absolute difference between the class probabilities, reflecting the inherent complexity from the model's perspective.
}
    \label{fig:pipeline:score}
\end{figure}

\section{Methodology}\seclabel{method}

We propose a self-adaptive curriculum learning paradigm that relies on the difficulty predicted by the PLM itself.
We use prompt templates (cf. \secref{template}) and the verbalizer component (cf. \secref{verbalizer}) to obtain the class probabilities, based on which we compute the difficulty score for each example (cf. \secref{difficulty}).
With the scores, we propose different sampling strategies for fine-tuning (cf. \secref{sampling}).

\subsection{Prompt Construction}\seclabel{template}

Our approach begins with the construction of task-specific prompts. 
The general structure is: 
\begin{equation*}
\texttt{Text} + \texttt{Template} 
\end{equation*}
where $\texttt{Text}$ is the actual text for which we want to obtain a prediction and $\texttt{Template}$ is a few tokens that help the model to understand the task and make a prediction. 
$\texttt{Template}$ always contains a special token \texttt{[MASK]}. 
We check the token distribution over vocabularies at the \texttt{[MASK]} position.

For example, in a sentiment analysis task for movie reviews, the prompt is formulated as: ``This is a very ambitious project, this was a \texttt{[MASK]} movie.'', where the first half, i.e., ``This is a very ambitious project'' is the actual sentence for classification while the rest is the template.
Here, \texttt{[MASK]} prompts the model to predict an adjective token (e.g., \textit{great}, \textit{bad}), reflecting the sentiment of a ``reviewer''. 
The prompt templates we use for each downstream task are shown in \secref{training_details}.

\subsection{Verbalizer Design}\seclabel{verbalizer}

A verbalizer maps the token predicted at the \texttt{[MASK]} position to a task-specific category label. 
Taking binary classification for example, we define the verbalizer with carefully selected keywords aligned with the dataset and the task context:
\begin{align*}
V = \{ & \text{positive} \rightarrow \texttt{positive keyword}, \notag \\
       & \text{negative} \rightarrow \texttt{negative keyword} \}
\end{align*}
where positive/negative refer to the category, and \texttt{positive/negative keyword} are the tokens we use representing the corresponding category.
Although multiple keywords per class can be considered, both previous research \citep{ma-etal-2023-prompt} and our preliminary results indicate that optimal performance is achieved when mapping each category to a single, clearly representative keyword. 
This verbalizer design is easily extendable to multi-class scenarios.
We show our verbalizers in \secref{training_details}.

\subsection{Difficulty Score Calculation}\seclabel{difficulty}

By feeding a prompt, we check the model's output logits at the \texttt{[MASK]} position. 
For each token $w_i$ in the vocabulary $\mathbb{V}$, we obtain its corresponding logit $z_i$.
We then calculate the probability of the token with the softmax function:
$
P(w_i) = \frac{e^{z_i}}{\sum_{w_j \in \mathbb{V}} e^{z_j}}
$

Then, we extract the \textbf{label-specific probabilities} using verbalizers.
Taking sentiment analysis (a \textbf{binary classification} task, for example, we compute the class probability by considering the selected keyword for each class:
\begin{align*}
P_{\text{pos}} &= P(\texttt{positive keyword}) \notag \\
P_{\text{neg}} &= P(\texttt{negative keyword})
\end{align*}

Note that $P_{\text{pos}}$ and $P_{\text{neg}}$ are normalized so that $P_{\text{pos}} + P_{\text{neg}} = 1 $.
The difficulty score is then defined as the absolute difference between the two class probabilities:
$
\text{Difficulty Score} = |P_{\text{pos}} - P_{\text{neg}}|
$.

Figure \ref{fig:pipeline:score} illustrates the process of calculating the difficulty score.
\textbf{The intuition is that a higher score indicates greater model confidence (lower difficulty), whereas a lower score suggests uncertainty (higher difficulty).} 
Our empirical results verify this intuition: Figure \ref{fig:scores} shows that, even before training, examples with higher scores (less difficult) generally correspond to correct predictions.
After training, the distribution shifts significantly toward higher scores (many examples become less difficult because the model has seen them), validating the effectiveness of our difficulty scoring method.
This method easily generalizes to \textbf{multi-class classification} by defining difficulty score as the margin between the two highest class probabilities:
$
\text{Difficulty Score} = |P_{\max} - P_{\text{second-max}}|
$.

\begin{figure*}
    \centering
    \setlength{\abovecaptionskip}{-0.01cm}
    \includegraphics[width=0.9\textwidth]{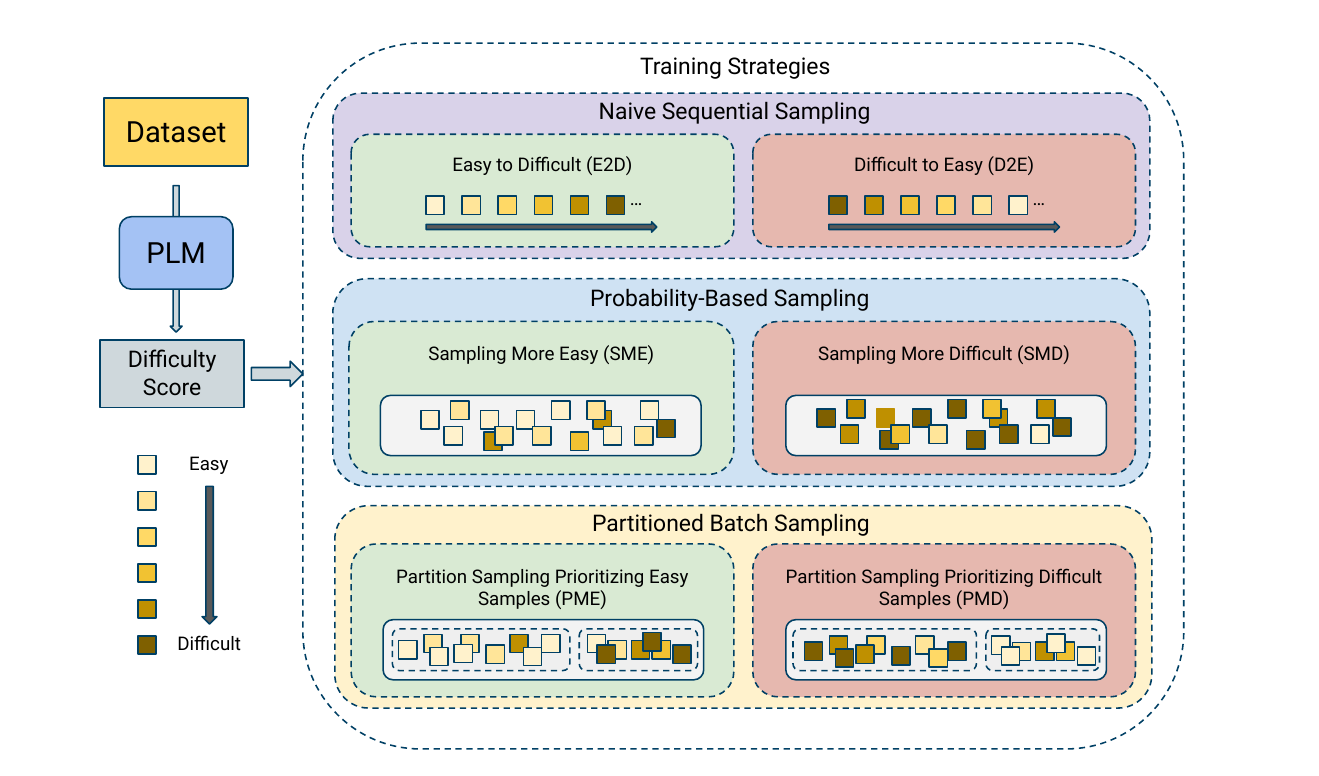}
    \caption{An illustration of our sampling strategies. Each example is associated with a difficulty score based on the PLM itself. Six sampling strategies are presented: \textbf{Naive Sequential Sampling} (E2D and D2E), \textbf{Probability-Based Sampling} (SME and SMD), and \textbf{Partitioned Batch Sampling} (PME and PMD). The difficulty of examples is indicated by color, with lighter colors representing easier samples and darker colors representing more difficult ones.
}
    \label{fig:pipeline}
\end{figure*}

\subsection{Sampling Strategies}\seclabel{sampling}

Drawing inspiration from curriculum learning, we propose six sampling strategies grouped into three categories. The sampling relies on the difficulty score of each example.
These strategies are designed to prioritize \textbf{``worth-learning''} examples during fine-tuning for specific tasks.  Figure~\ref{fig:pipeline} presents an overview of our sampling strategies. 

\subsubsection{Naive Sequential Sampling}

The most straightforward approach, akin to curriculum learning, is to arrange the training examples based on their difficulty scores and train the model using a fixed order.
Let \( X = \{x_n\}_{n=1}^{N} \) be the training examples, sorted by their associated difficulty scores $s_n$ in \textbf{either ascending or descending order}.
We propose two sampling strategies.

\paragraph{Easy to Difficult (E2D)} Training examples are sorted \textbf{descendingly} according to the scores, such that $s_1 \geq s_2 \geq \dots \geq s_n$, with $x_1$ being the easiest one and $x_n$ the hardest one.
Models are exposed to examples from $x_1$ to $x_N$ sequentially.

\paragraph{Difficult to Easy (D2E)} Training examples are sorted \textbf{ascendingly} according to the scores, such that $s_1 \leq s_2 \leq \dots \leq s_n$,  with $x_1$ being the hardest one and $x_n$ the easiest one.
Models are exposed to examples from $x_1$ to $x_N$ sequentially.

\subsubsection{Probability-Based Sampling}

Our intuition is that sequentially exposing examples to the model can be overly rigid and lack diversity. This might result in the model's degradation in dealing with very easy or difficult examples.
Therefore, we propose probability-based sampling strategies that introduce a more flexible and diverse training flow. 
Specifically, rather than following a fixed order, \textbf{examples are assigned probabilities based on their difficulty rankings}, enabling the model to encounter a controlled mixture of easy and hard examples.
Given the ordered examples $ X = [x_1, x_2, ..., x_N]$ according to their scores, the sampling probability for $x_n$ is defined as:
\begin{equation*}
P(x_n) = \frac{n^2}{\sum_{j=1}^{N} j^2}
\end{equation*}
That is, the sampling probability from $x_1$ to $x_N$ increases. 
We propose two sampling strategies.

\paragraph{Sampling More Easy (SME)} Training examples are sorted \textbf{ascendingly} according to the scores; thus, easier examples (higher ranks $n$) have larger probabilities of being sampled. This results in a sampling behavior in favor of easy examples with occasional difficult ones.

\paragraph{Sampling More Difficult (SMD)}
Training examples are sorted in \textbf{descending} order according to the scores; thus, more difficult examples (higher ranks $n$) have larger probabilities of being sampled. This results in a sampling behavior in favor of hard examples with occasional easy ones.

\subsubsection{Partitioned Batch Sampling Strategies}

As an extension of probability-based sampling, this method allows fine-grained control within each batch.
Each batch $B$ contains two partitions ($B_1$ and $B_2$) of examples, with \textbf{one partition focusing on sampling easier examples, while the other on more difficult ones}.
Note that sampling within each partition is still based on the probability, rather than being deterministic.
This also ensures diversity and avoids overfitting to a fixed progression. 
This approach enables a more dynamic and balanced mixture of easy and hard samples during fine-tuning. 
\textbf{We set $|B_1| > |B_2|$, aiming to give higher priority to partition $B_1$ during fine-tuning}.\footnote{We set $|B_1|:|B_2|=6:4$ based on preliminary results.} 
We propose two sampling strategies.

\paragraph{Prioritizing Easy Samples (PME)}
The first partition \( B_1 \) prioritizes easy samples, while the second partition \( B_2 \) prioritizes difficult examples, achieved by assigning \textbf{two different probabilities to each example $x_n$}, one for $B_1$ and the other for $B_2$:
\begin{equation*}
P_{B_1}(x_n) = \frac{n^2}{\sum_{j=1}^{N} j^2}, \quad P_{B_2}(x_n) = \frac{(N-n)^2}{\sum_{j=1}^{N} j^2}
\end{equation*}
In PME, the training examples are sorted in \textbf{ascending order} according to the scores.
In this way, $P_{B_1}(x_n)$ prioritizes on easier examples while $P_{B_2}(x_n)$ prioritizes on harder examples.

\paragraph{Prioritizing Difficult Samples (PMD)}
Conversely, the training examples are sorted in \textbf{descending order} according to the scores.
In this way, $P_{B_1}(x_n)$ prioritizes on harder examples while $P_{B_2}(x_n)$ prioritizes on easier examples.

\begin{table*}[ht]
\footnotesize
\centering
\setlength{\tabcolsep}{1.4mm}
\definecolor{softred}{RGB}{255, 230, 230}
\definecolor{softblue}{RGB}{220, 230, 255}
\resizebox{\textwidth}{!}{%
\begin{tabular}{llcccccccccccccccc}
\toprule
& & \multicolumn{4}{c}{SST-2} & \multicolumn{4}{c}{SST-5} & \multicolumn{4}{c}{HSOL} & \multicolumn{4}{c}{XNLI} \\
\cmidrule(lr){3-6} \cmidrule(lr){7-10} \cmidrule(lr){11-14} \cmidrule(lr){15-18}
& & Acc & F1 & Prec & Rec & Acc & F1 & Prec & Rec & Acc & F1 & Prec & Rec & Acc & F1 & Prec & Rec \\
\midrule
\multirow{8}{*}{BERT}
  & Random & 91.97 & 91.97 & 91.99 & 91.96 & \underline{53.62} & \textbf{52.37} & 53.18 & \textbf{52.05} & \underline{91.67} & 73.58 & \underline{80.15} & 71.76 & \textbf{84.01} & \textbf{84.02} & \textbf{84.23} & \textbf{84.01} \\
  & Length & 92.09 & 92.08 & 92.15 & 92.06 & 51.75 & 51.03 & 51.52 & \underline{51.70} & 91.50 & 70.99 & \textbf{80.30} & 69.04 & 83.06 & 83.07 & 83.22 & 83.06 \\
  & E2D & \cellcolor{softred}\underline{92.39} & \cellcolor{softred}\underline{92.39} & \cellcolor{softred}\underline{92.39} & \cellcolor{softred}\underline{92.41} & 52.16 & \cellcolor{softblue}47.12 & \cellcolor{softred}\textbf{56.88} & \cellcolor{softblue}48.17 & \cellcolor{softblue}90.95 & \cellcolor{softblue}70.20 & \cellcolor{softblue}76.21 & 70.50 & \cellcolor{softblue}82.83 & \cellcolor{softblue}82.87 & 83.52 & \cellcolor{softblue}82.83 \\
  & D2E & \cellcolor{softblue}91.93 & \cellcolor{softblue}91.92 & 92.12 & \cellcolor{softblue}91.88 & \cellcolor{softblue}51.60 & \cellcolor{softblue}50.48 & 52.40 & \cellcolor{softblue}50.01 & \cellcolor{softblue}91.23 & \cellcolor{softred}74.23 & \cellcolor{softblue}77.04 & \cellcolor{softred}72.81 & \cellcolor{softblue}82.12 & \cellcolor{softblue}82.24 & 83.23 & \cellcolor{softblue}82.12 \\
  & SME & \cellcolor{softblue}91.25 & \cellcolor{softblue}91.23 & \cellcolor{softblue}91.35 & \cellcolor{softblue}91.20 & 52.91 & \cellcolor{softblue}49.78 & \cellcolor{softred}53.71 & \cellcolor{softblue}50.39 & \cellcolor{softred}\textbf{91.81} & \cellcolor{softred}73.83 & \cellcolor{softblue}79.76 & \cellcolor{softred}72.88 & 83.08 & 83.10 & 83.73 & 83.08 \\
  & SMD & \cellcolor{softblue}91.48 & \cellcolor{softblue}91.47 & \cellcolor{softblue}91.53 & \cellcolor{softblue}91.45 & 52.73 & \cellcolor{softblue}50.92 & 51.84 & \cellcolor{softblue}51.14 & 91.51 & \cellcolor{softred}\underline{74.71} & \cellcolor{softblue}79.21 & \cellcolor{softred}72.22 & \cellcolor{softblue}82.31 & \cellcolor{softblue}82.41 & 83.28 & \cellcolor{softblue}82.31 \\
  & PME & \cellcolor{softblue}91.40 & \cellcolor{softblue}91.38 & \cellcolor{softblue}91.59 & \cellcolor{softblue}91.35 & \cellcolor{softred}\textbf{53.83} & \cellcolor{softblue}50.72 & \cellcolor{softred}\underline{54.33} & \cellcolor{softblue}50.40 & \underline{91.67} & \cellcolor{softred}74.46 & \cellcolor{softblue}79.19 & \cellcolor{softred}\underline{73.05} & \underline{83.75} & \underline{83.78} & \underline{84.02} & \underline{83.75} \\
  & PMD & \cellcolor{softred}\textbf{92.62} & \cellcolor{softred}\textbf{92.61} & \cellcolor{softred}\textbf{92.73} & \cellcolor{softred}\textbf{92.60} & 52.73 & \underline{51.66} & \cellcolor{softred}53.56 & \cellcolor{softblue}51.59 & 91.64 & \cellcolor{softred}\textbf{76.14} & \cellcolor{softblue}78.43 & \cellcolor{softred}\textbf{74.76} & 83.27 & 83.29 & 83.54 & 83.27 \\
\midrule
\multirow{8}{*}{RoBERTa}
  & Random & \textbf{94.11} & \textbf{94.11} & 94.15 & \textbf{94.10} & 56.00 & \underline{54.34} & 56.55 & 54.62 & \underline{92.18} & 75.27 & 81.79 & 72.76 & 87.11 & 87.11 & 87.28 & 87.11 \\
  & Length & 93.35 & 93.34 & 93.46 & 93.31 & 54.27 & 53.17 & 52.92 & \underline{54.95} & 92.00 & 67.41 & \textbf{85.02} & 65.60 & 86.20 & 86.14 & 86.37 & 86.20 \\
  & E2D & \underline{93.92} & \underline{93.92} & \cellcolor{softred}\textbf{95.95} & \underline{93.91} & \cellcolor{softred}\underline{57.00} & 53.29 & \cellcolor{softred}56.64 & \cellcolor{softblue}53.76 & \cellcolor{softblue}90.96 & 73.98 & \cellcolor{softblue}77.04 & \cellcolor{softred}74.38 & \cellcolor{softblue}85.73 & \cellcolor{softblue}85.76 & \cellcolor{softblue}86.23 & \cellcolor{softblue}85.73 \\
  & D2E & 93.54 & 93.54 & 93.57 & 93.52 & \cellcolor{softred}\textbf{57.07} & \cellcolor{softred}\textbf{55.30} & 56.00 & \cellcolor{softred}\textbf{55.70} & \cellcolor{softblue}91.43 & 73.66 & \cellcolor{softblue}79.06 & 71.85 & \cellcolor{softred}\underline{87.39} & \cellcolor{softred}\underline{87.43} & \cellcolor{softred}\underline{87.57} & \cellcolor{softred}\underline{87.39} \\
  & SME & 93.35 & 93.34 & \cellcolor{softred}\underline{94.44} & 93.33 & 55.49 & \cellcolor{softblue}50.76 & \cellcolor{softred}\textbf{57.76} & \cellcolor{softblue}51.11 & \cellcolor{softblue}91.76 & \cellcolor{softred}\underline{75.46} & \cellcolor{softblue}79.36 & \cellcolor{softred}\textbf{75.79} & 87.11 & \cellcolor{softred}87.13 & 87.25 & 87.11 \\
  & SMD & 93.39 & 93.37 & 93.56 & 93.34 & \cellcolor{softred}56.46 & 53.83 & 56.50 & \cellcolor{softblue}53.51 & \cellcolor{softblue}91.57 & 75.23 & \cellcolor{softblue}78.14 & \cellcolor{softred}74.09 & 86.86 & 86.96 & \cellcolor{softred}87.42 & 86.86 \\
  & PME & 93.85 & 93.84 & 93.89 & 93.82 & 55.76 & \cellcolor{softblue}52.17 & \cellcolor{softred}57.13 & \cellcolor{softblue}52.64 & 92.14 & \cellcolor{softred}\textbf{77.27} & \cellcolor{softblue}80.05 & \cellcolor{softred}\underline{75.64} & 86.86 & 86.91 & 87.17 & 86.86 \\
  & PMD & \cellcolor{softblue}93.27 & \cellcolor{softblue}93.27 & \cellcolor{softblue}93.36 & \cellcolor{softblue}93.27 & \cellcolor{softred}56.89 & 54.15 & \cellcolor{softred}\underline{57.22} & \cellcolor{softblue}54.04 & \cellcolor{softred}\textbf{92.53} & 74.96 & \underline{82.89} & \cellcolor{softred}73.71 & \cellcolor{softred}\textbf{87.47} & \cellcolor{softred}\textbf{87.49} & \cellcolor{softred}\textbf{87.58} & \cellcolor{softred}\textbf{87.47} \\
\bottomrule
\end{tabular}
}
\caption{
Comparison of different sampling strategies and baselines across four datasets (SST-2, SST-5, HSOL, and XNLI) using BERT and RoBERTa as backbone models. 
Accuracy, F1 score, precision, and recall are reported. 
\textbf{Bold} (resp. \underline{underlined}) entries highlight the best (resp. second-best) performance within each model group.
For our proposed sampling approaches, we additionally use background colors
\textcolor{softred}{red} to indicate values higher than both baselines, \textcolor{softblue}{blue} to indicate values lower than both, and white to indicate performance between the two baselines. All results are averaged over runs with 3 different random seeds. 
}

\label{tab:general_result}
\end{table*}

\section{Experimental Setup}

We evaluate our proposed methods on four publicly available datasets, covering diverse NLP tasks to demonstrate the generality of our approach. 

\subsection{Datasets}

\paragraph{Stanford Sentiment Treebank Binary (SST-2)}

SST-2~\citep{socher-etal-2013-recursive} is a balanced binary sentiment analysis dataset containing movie review sentences labeled as positive or negative.

\paragraph{Fine-grained Sentiment Analysis (SST-5)} 
    
SST-5 dataset~\citep{socher-etal-2013-recursive} contains sentences from movie reviews labeled into five fine-grained sentiment categories: very positive, positive, neutral, negative, and very negative.

\paragraph{Hate Speech Offensive Language (HSOL)} 
    
The Hate Speech Offensive Language dataset~\citep{hateoffensive} includes tweets labeled into three categories: hate speech, offensive language, and neither, with a significant class imbalance.

\paragraph{Cross-lingual Natural Language Inference (XNLI)} 
    
XNLI~\citep{conneau2018xnli} is a widely-used benchmark for natural language understanding tasks, providing sentence pairs labeled in three categories: entailment, neutral, or contradiction.

\subsection{Models}

We use \texttt{bert-base-uncased} (BERT-base) \citep{devlin2018bert} and \texttt{roberta-base} (RoBERTa-base)~\citep{liu2019robertarobustlyoptimizedbert} as the base PLMs for all experiments.
Since masked language modeling is the main objective in their pretraining,
both models have a special \texttt{[MASK]} token in their vocabularies, which allows us to compute the difficulty score for each example in the training set of the downstream dataset and apply our sampling strategy for prompt-based fine-tuning, as introduced in \secref{method}.

\subsection{Baselines}

We consider two baselines: \textbf{Random} and \textbf{Length}.
The \textbf{Random} baseline follows the classic strategy where a batch of training examples is randomly sampled from the training dataset.
The \textbf{Length} baseline assumes that examples with more tokens are more difficult \citep{platanios-etal-2019-competence}.
The examples are sorted from shortest to longest according to their tokenized length. 
\textbf{Length} not only reflects the inherent sentence length but also captures word rarity, as rare or uncommon words are typically tokenized into multiple subword units, thus resulting in longer sequences.

\section{Results and Discussions}

\subsection{Main Result}

Table \ref{tab:general_result} presents the accuracy, F1 score, precision, and recall scores on the test sets of the 4 datasets from the baselines and our training strategies.

\paragraph{RoBERTa consistently outperforms BERT across all datasets.}
RoBERTa shows overall better performance than BERT across all datasets under almost all sampling strategies, including Random and Length baseline.
This is a strong indicator that RoBERTa's pretrained representation provides stronger generalization, especially under low-resource or imbalanced sampling conditions.

\paragraph{Random sampling is occasionally sufficient, but curriculum-sampling strategies offer more robust improvements.} 
While the baseline Random shows fair performance, especially in low-difficulty or well-balanced datasets (like SST-2), it gains inconsistent performance across harder datasets like SST-5 and XNLI.
The baseline Length achieves slightly better performance than Random, indicating that curriculum learning with the sentence length as an indicator of difficulty works.
However, the performance is also less consistent and usually worse than our proposed approaches.
Our sampling strategies, especially PME, E2D, and SME, tend to offer more consistent gains, indicating the effectiveness of using the model's own prediction for difficulty calculation of training examples.

\paragraph{PMD achieves the highest performance in most cases.}
The PMD strategy yields top performance (highlighted in bold) on multiple datasets for both BERT and RoBERTa, especially on SST-2 and XNLI. Its consistent superiority suggests that its dynamic sampling mechanism effectively emphasizes worth-learning examples during training.

\paragraph{Dataset difficulty affects the benefit of sampling strategies.}
On easier datasets such as SST-2 and HSOL, most strategies achieve high and stable results, and the performance gap between baselines and sampling-based methods remains relatively small. 
In contrast, on more challenging datasets like SST-5 and XNLI, the performance differences are more pronounced, indicating that sampling strategies provide greater benefits when the task involves finer-grained classes.

\paragraph{On imbalanced datasets, the proposed sampling strategies offer clear advantages.}
In datasets like HSOL, which exhibit label imbalance or fine-grained distinctions, our sampling strategies, such as PME and SME, consistently achieve higher F1 scores compared to baselines. 
This indicates their effectiveness in promoting better representation of minority or harder-to-learn classes, improving the overall balance between precision and recall.

\subsection{Training Progression Analysis}

To further understand the benefit of our methods, we analyze the changes in accuracy and loss on the validation set for each dataset within a single epoch of fine-tuning. 
Throughout the epoch, we store a checkpoint every 10\% of the training samples. 
We then evaluate each checkpoint on the validation set.
Consequently, we save the average accuracy and loss on the validation set at 10 different checkpoints.
We discuss the trend of accuracy of SST-2 and SST-5 in the following. 
The complete results (accuracy and loss) for each dataset are presented in \secref{result_details}.

\begin{figure} 
    \centering
    \includegraphics[width=0.46\textwidth]{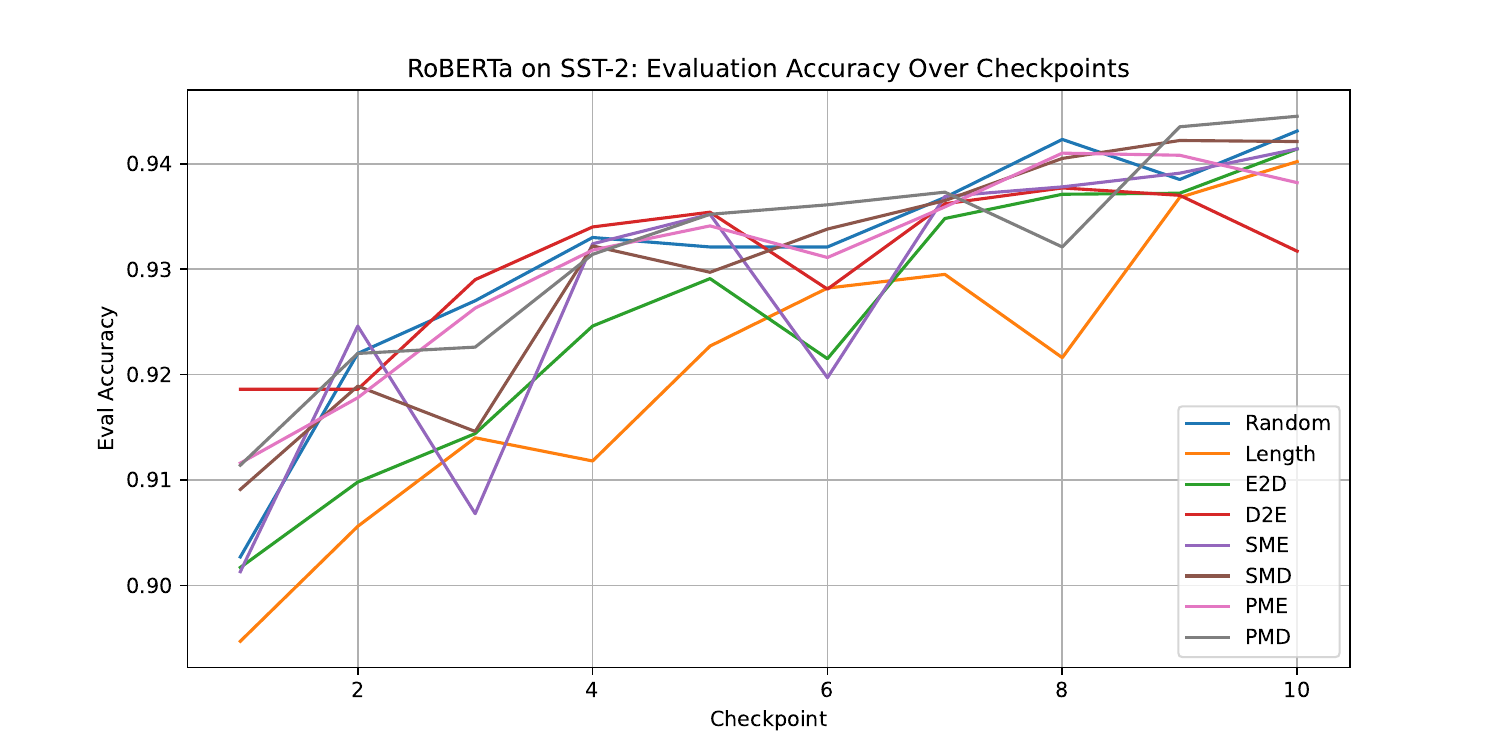}
    \caption{Progression of accuracy during a single epoch on \textbf{SST-2}. 
    Each checkpoint corresponds to a model seeing 10\% of the training examples.
}
\label{fig:robertasst2acc}
\end{figure}

Figure \ref{fig:robertasst2acc} presents the RoBERTa results on \textbf{SST-2}.
At the first checkpoint, sampling strategies D2E, PME, PMD, and SMD show a clear advantage, far exceeding both the baselines and the E2D and SME strategies. 
This might indicate that early exposure to difficult examples might be helpful.
Throughout training, all methods exhibit some degree of fluctuation.
At the final checkpoint, most methods, including the baselines, continue to improve.
This suggests that, despite fluctuations during training, most methods benefit from longer training time.
\begin{figure}[H]   
    \centering
    \includegraphics[width=0.46\textwidth]
    {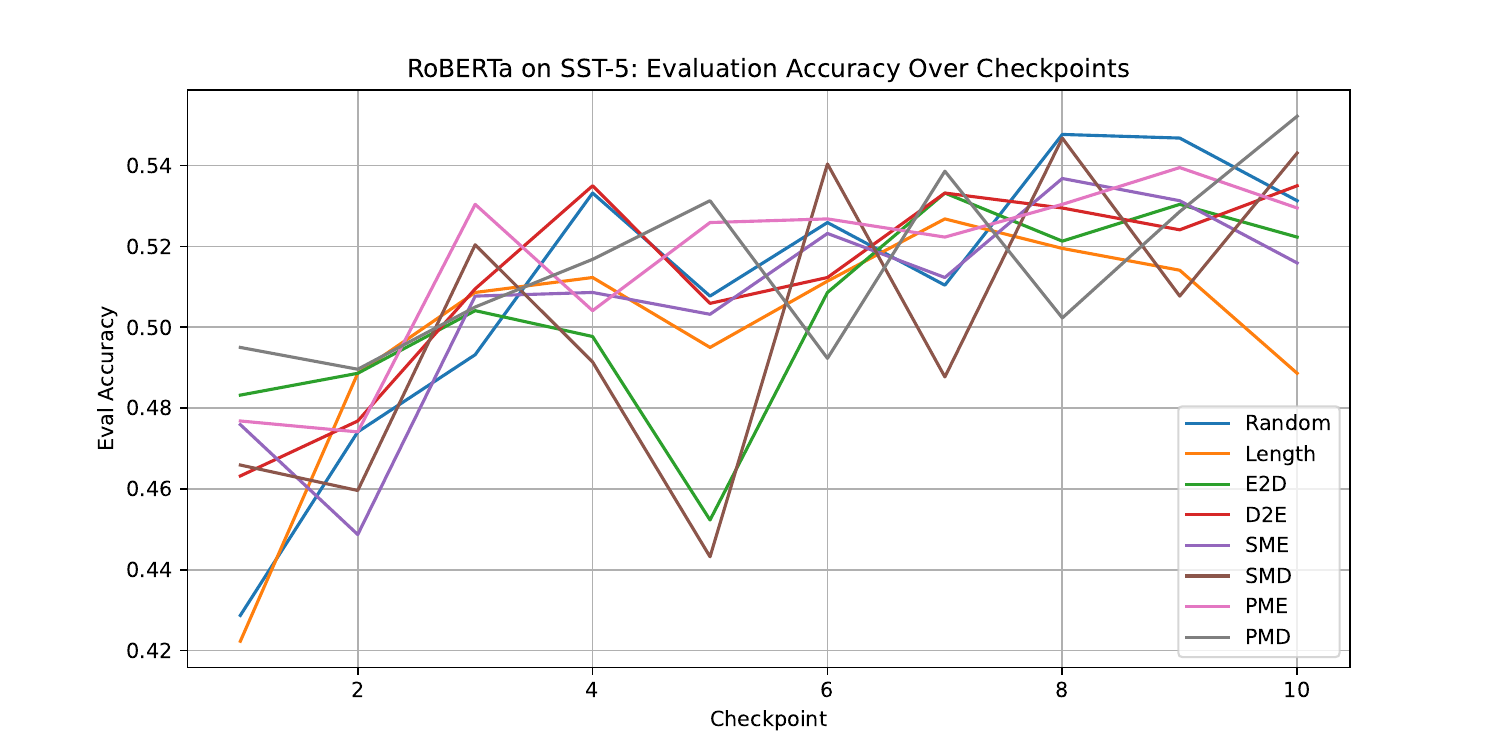}
    \caption{Progression of accuracy during a single epoch on \textbf{SST-5}. 
    Each checkpoint corresponds to a model seeing 10\% of the training examples.
}
    \label{fig:robertasst5acc}
\end{figure}

Figure \ref{fig:robertasst5acc} presents the RoBERTa results on \textbf{SST-5}. Different from the trend from SST-2, we observe that all our training strategies significantly outperform the baseline at the first checkpoint, indicating that, compared to Random and Length, our methods enable RoBERTa to learn useful features more rapidly in the early stages.
However, almost all strategies exhibit substantial fluctuations throughout training. 
In the final phase, PMD, SMD, and D2E still show improvements, while other strategies decline. 
Among them, PMD achieves the highest performance through a rapid increase. 
This might suggest that, for multi-class classification, prioritizing difficult samples can facilitate more stable learning in the last stage of training.

\begin{table*}[ht]
\footnotesize
\centering
\setlength{\tabcolsep}{1.4mm}
\definecolor{softred}{RGB}{255, 230, 230}
\definecolor{softblue}{RGB}{220, 230, 255}
\resizebox{\textwidth}{!}{%
\begin{tabular}{llcccccccccccccccc}
\toprule
& & \multicolumn{4}{c}{SST-2} & \multicolumn{4}{c}{SST-5} & \multicolumn{4}{c}{HSOL} & \multicolumn{4}{c}{XNLI} \\
\cmidrule(lr){3-6} \cmidrule(lr){7-10} \cmidrule(lr){11-14} \cmidrule(lr){15-18}
& & Acc & F1 & Prec & Rec & Acc & F1 & Prec & Rec & Acc & F1 & Prec & Rec & Acc & F1 & Prec & Rec \\
\midrule
\multirow{8}{*}{BERT}
  & Random & 80.85 & 80.60 & \underline{81.92} & 80.78 & 36.58 & 31.22 & 38.43 & 34.07 & 77.69 & 32.58 & \textbf{47.70} & 34.96 & 35.52 & 33.41 & 36.07 & 35.52 \\
  & Length & 68.12 & 65.18 & 75.96 & 67.62 & 37.36 & \underline{32.68} & 35.87 & \textbf{35.19} & 77.29 & \textbf{39.01} & 42.80 & \textbf{39.05} & 35.07 & 25.99 & \textbf{37.38} & 35.07 \\
  & E2D & \cellcolor{softblue}52.29 & \cellcolor{softblue}36.97 & \cellcolor{softblue}72.21 & \cellcolor{softblue}51.41 & \cellcolor{softblue}30.94 & \cellcolor{softblue}24.97 & \cellcolor{softblue}29.59 & 33.08 & \cellcolor{softblue}76.63 & \underline{37.56} & 45.87 & 37.47 & \cellcolor{softblue}34.32 & 30.89 & \cellcolor{softblue}35.33 & \cellcolor{softblue}34.32 \\
  & D2E & \cellcolor{softred}\textbf{84.02} & \cellcolor{softred}\textbf{84.01} & \cellcolor{softred}\textbf{84.05} & \cellcolor{softred}\textbf{84.01} & \cellcolor{softred}39.23 & \cellcolor{softblue}24.28 & \cellcolor{softred}\textbf{45.59} & \cellcolor{softblue}31.31 & \cellcolor{softblue}76.28 & 34.83 & \cellcolor{softblue}41.11 & 35.89 & \cellcolor{softblue}33.73 & 27.70 & \cellcolor{softblue}35.31 & \cellcolor{softblue}33.73 \\
  & SME & \cellcolor{softred}\underline{81.00} & \cellcolor{softred}\underline{80.94} & 81.34 & \cellcolor{softred}\underline{80.99} & \cellcolor{softred}\textbf{40.44} & \cellcolor{softblue}30.22 & \cellcolor{softred}44.68 & \cellcolor{softblue}33.88 & \cellcolor{softblue}77.07 & 37.34 & 45.87 & \underline{37.66} & 35.24 & 31.68 & \cellcolor{softblue}35.77 & 35.24 \\
  & SMD & 78.36 & 78.13 & 79.36 & 78.35 & \cellcolor{softred}\textbf{40.44} & \cellcolor{softblue}29.60 & \cellcolor{softred}39.47 & \cellcolor{softblue}33.44 & \cellcolor{softblue}77.28 & 34.42 & \underline{47.53} & 36.05 & 35.32 & 29.17 & \cellcolor{softblue}36.04 & 35.32 \\
  & PME & 79.70 & 79.49 & 80.71 & 78.35 & \cellcolor{softred}\underline{39.67} & 32.39 & 37.98 & 34.42 & \cellcolor{softred}\underline{77.89} & 35.22 & 47.04 & 36.48 & \cellcolor{softred}\underline{36.10} & \cellcolor{softred}\underline{35.20} & 36.21 & \cellcolor{softred}\underline{36.10} \\
  & PMD & 79.32 & 78.87 & 81.38 & 79.31 & \cellcolor{softred}40.62 & \cellcolor{softred}32.91 & 38.29 & 35.13 & \cellcolor{softred}\textbf{77.92} & 34.91 & 45.60 & 36.44 & \cellcolor{softred}\textbf{36.39} & \cellcolor{softred}\textbf{35.69} & \underline{36.40} & \cellcolor{softred}\textbf{36.39} \\
\midrule
\multirow{8}{*}{RoBERTa}
  & Random & 89.64 & 89.63 & 89.75 & 89.66 & 45.11 & \underline{34.54} & 44.46 & 38.21 & \textbf{80.67} & \textbf{42.44} & 53.47 & \underline{41.42} & 35.53 & \textbf{31.59} & 33.70 & 35.53 \\
  & Length & 84.02 & 83.77 & 85.39 & 83.85 & 40.77 & 28.69 & 40.64 & 33.25 & 79.35 & 39.33 & 50.57 & 39.22 & 35.26 & 27.62 & 26.37 & 35.26 \\
  & E2D & 87.99 & 87.99 & 88.02 & 87.99 & 43.18 & \cellcolor{softred}\textbf{35.35} & \cellcolor{softblue}39.61 & 37.51 & \cellcolor{softblue}77.20 & \cellcolor{softblue}35.50 & \cellcolor{softblue}50.24 & \cellcolor{softblue}36.20 & \cellcolor{softred}\underline{35.55} & 29.81 & \cellcolor{softred}34.73 & \cellcolor{softred}\underline{35.55} \\
  & D2E & \cellcolor{softred}\underline{90.56} & \cellcolor{softred}\underline{90.55} & \cellcolor{softred}\underline{90.58} & \cellcolor{softred}\underline{90.54} & 45.10 & \cellcolor{softblue}26.26 & \cellcolor{softblue}34.92 & 35.65 & \cellcolor{softblue}77.69 & \cellcolor{softblue}38.49 & \cellcolor{softblue}47.48 & \cellcolor{softblue}38.63 & \cellcolor{softblue}32.48 & \cellcolor{softblue}22.15 & 32.98 & \cellcolor{softblue}32.48 \\
  & SME & \cellcolor{softred}90.37 & \cellcolor{softred}90.36 & \cellcolor{softred}90.42 & \cellcolor{softred}90.34 & \cellcolor{softred}\underline{45.69} & 34.29 & \cellcolor{softblue}39.87 & \cellcolor{softred}\textbf{38.32} & \cellcolor{softblue}78.62 & \cellcolor{softblue}34.20 & \cellcolor{softred}\textbf{56.91} & \cellcolor{softblue}36.01 & \cellcolor{softred}\textbf{35.61} & 29.38 & \cellcolor{softred}\underline{34.86} & \cellcolor{softred}\textbf{35.61} \\
  & SMD & \cellcolor{softred}\textbf{90.86} & \cellcolor{softred}\textbf{90.86} & \cellcolor{softred}\textbf{90.87} & \cellcolor{softred}\textbf{90.86} & 43.79 & \cellcolor{softblue}28.46 & \cellcolor{softblue}37.17 & 35.63 & \cellcolor{softblue}78.68 & \cellcolor{softblue}35.17 & \cellcolor{softred}\underline{54.04} & \cellcolor{softblue}36.62 & \cellcolor{softblue}33.27 & 28.36 & \cellcolor{softred}\underline{34.86} & \cellcolor{softblue}33.27 \\
  & PME & 88.95 & 88.93 & 89.17 & 88.93 & \cellcolor{softred}\textbf{47.41} & 31.39 & \cellcolor{softred}\underline{44.86} & \cellcolor{softred}\underline{38.24} & \underline{80.64} & \underline{42.04} & \cellcolor{softred}53.97 & \cellcolor{softred}\textbf{41.76} & \cellcolor{softblue}34.29 & 30.90 & \cellcolor{softred}\textbf{35.02} & \cellcolor{softblue}34.29 \\
  & PMD & \cellcolor{softred}90.06 & \cellcolor{softred}90.03 & \cellcolor{softred}90.35 & \cellcolor{softred}90.01 & \cellcolor{softred}45.13 & 32.33 & \cellcolor{softred}\textbf{45.13} & 37.11 & 79.99 & 41.34 & 52.05 & 40.84 & \cellcolor{softblue}33.83 & \underline{31.13} & \cellcolor{softred}34.57 & \cellcolor{softblue}33.83 \\
\bottomrule
\end{tabular}
}
\caption{
Comparison of different sampling strategies and baselines across four datasets (SST-2, SST-5, HSOL, and XNLI) under \textbf{few-shot learning} setting with 64 training instances.
Accuracy, F1 score, precision, and recall are reported. 
\textbf{Bold} (resp. \underline{underlined}) entries highlight the best (resp. second-best) performance within each model group.
For our proposed sampling approaches, we additionally use background colors
\textcolor{softred}{red} to indicate values higher than both baselines, \textcolor{softblue}{blue} to indicate values lower than both, and white to indicate performance between the two baselines. All results are averaged over runs with 3 different random seeds.
}

\label{tab:fewshot_result}
\end{table*}

\subsection{Few-Shot Learning}

To further investigate the benefit of our strategies under the scenarios where limited training data are present, we conduct a few-shot learning evaluation, similar to the setup of \citet{ma-etal-2023-prompt}, using the 4 datasets. Specifically, we select the \textbf{top 64} ranked examples in each sampling strategy.\footnote{We use the top 64 ranked examples for all strategies except Random, for which examples are randomly sampled.} 
The number of 64 samples is chosen to ensure 
sufficient diversity across difficulty levels.
The PLMs are trained on these examples solely, and Table \ref{tab:fewshot_result} presents the results of the resulting models on the test set.

\paragraph{RoBERTa shows a clear advantage over BERT, especially on SST-2 and SST-5.} 
Similar to the results shown in Table \ref{tab:general_result}, RoBERTa also achieves better performance than BERT. 
We even notice that the performance on SST-2 is already close to the fully supervised performance reported in Table \ref{tab:general_result}.
For HSOL and XNLI, however, the gap between the two models is much smaller.
We assume this is due to dataset imbalance and difficulty, which limit the effectiveness of few-shot learning.

\paragraph{On SST-2 and SST-5, most of our sampling strategies consistently outperform both baselines except for E2D.}
Length performs noticeably worse than the other methods, which is because only short-length examples are exposed to the model.
On the other hand, the baseline Random remains relatively strong, as it sees both short and long examples.
We notice that E2D in BERT fails to train the model properly, which is expected since the model only sees easy examples on which the model should already perform very well, even without any fine-tuning.
For other training strategies, we generally see improvements.
Strategies such as D2E and probability-based methods like SME, PME, and PMD show substantial improvements across multiple metrics, indicating that hard examples are particularly important in few-shot learning.

\paragraph{For the more challenging inference dataset XNLI, using only 64 samples appears insufficient for training.} 
We notice that all models obtain much lower performance in XNLI compared with the results of full-dataset training (cf. Table \ref{tab:general_result}). 
This indicates the difficulty of XNLI dataset -- only when enough training instances are available, the model can learn the necessary features for making reasonable decisions.
As a result, based on the poor performance, it is difficult to draw clear conclusions regarding which sampling strategy is more effective on XNLI.
We hypothesize that increasing the number of training samples, e.g., 128 or 256, could alleviate the problem.

\section{Conclusion}

In this work, we introduced a self-adaptive curriculum learning paradigm that leverages a PLM's own confidence to estimate the difficulty of training examples. 
We further propose a range of sampling strategies: sequential, probabilistic, and partitioned, and verify the effectiveness on multiple NLU tasks. 
Our empirical results show improved performance in both full-data and few-shot settings, confirming the utility of model-predicted difficulty as a training signal. 
This paradigm offers a scalable and model-centric alternative to traditional curriculum learning, offering insights for broader applications across diverse NLU tasks.

\section*{Limitations}
We propose a self-adaptive curriculum learning paradigm that relies on the difficulty score predicted by the model itself.
Despite promising results, several limitations remain, particularly related to GPU memory constraints, which restrict input size and dataset coverage. 
With access to more powerful GPUs, we could conduct experiments on larger and more comprehensive datasets. We compare with representative baselines: \textbf{Random} and \textbf{Length}. Future work can also consider other difficulty-based alternatives, such as rarity- or attention-based sampling. 
Furthermore, our current experiments are limited to English classification tasks; future work should explore the applicability of our method to multilingual and cross-lingual settings.

Our current implementation is based on single-token classification settings. 
Extending difficulty scoring to multi-token or generative tasks (e.g., QA, summarization) remains an open direction. 
Furthermore, since prompt-based learning is highly sensitive to prompt design, experimenting with different templates and verbalizer words could further enhance model performance and interpretability. 
Another possible limitation is the lack of direct comparison with human-annotated difficulty levels, which could offer further insight into the alignment or divergence between model-based and human intuition.

Addressing imbalanced datasets by integrating dual curriculum learning concepts and implementing dynamic or multi-phase training strategies could also improve adaptability and efficiency. Overcoming these challenges would significantly boost the effectiveness and generalizability of our sampling strategies.

\section*{Acknowledgments}

This work was funded by Deutsche Forschungsgemeinschaft (project SCHU 2246/14-1).

\bibliography{custom}

\appendix

\section{Training Details}\seclabel{training_details}

We evaluate our proposed methods on four publicly available datasets, covering diverse NLP tasks to demonstrate the generality of our approach. Below we describe each dataset, including preprocessing, prompt templates, and verbalizer definitions.

\subsection{Stanford Sentiment Treebank Binary (SST-2)}
 
 We randomly partition the original training set into training (80\%) and validation sets (20\%), maintaining label distribution. The original validation set serves as our test set. Tokenized samples are truncated at 128 tokens. The prompt template and verbalizer are set as follows:
\[
x + \text{``this was a [MASK] movie.''}
\] 
\[
V = \{\text{positive}\rightarrow\text{``great''},\; \text{negative}\rightarrow\text{``bad''}\}
\]

\subsection{Fine-grained Sentiment Analysis (SST-5)}
 The maximum token length is set to 128 tokens. The prompt template and verbalizer are set as follows:
\[
x + \text{``this was a [MASK] movie.''}
\]
\[
V = \left\{
\begin{aligned}
    & \text{very positive} \rightarrow \text{``amazing''},\; \\
    & \text{positive} \rightarrow \text{``great''},\; \\
    & \text{neutral} \rightarrow \text{``okay''},\; \\
    & \text{negative} \rightarrow \text{``bad''},\; \\
    & \text{very negative} \rightarrow \text{``terrible''}
\end{aligned}
\right\}
\]

\subsection{Hate Speech Offensive Language (HSOL)}
 We split the original dataset into training (80\%), validation (10\%), and test (10\%) subsets, maintaining class distribution. Maximum token length is limited to 128 tokens. The prompt template and verbalizer are set as follows:
\[
x + \text{``this was [MASK].''}
\]
\[
V = \left\{
\begin{aligned}
    & \text{hate speech} \rightarrow \text{``hateful''},\; \\
    & \text{offensive} \rightarrow \text{``offensive''},\; \\
    & \text{neither} \rightarrow \text{``neutral''}
\end{aligned}
\right\}
\]
\subsection{Cross-lingual Natural Language Inference (XNLI)}
We limit maximum sequence length to 128 tokens. The prompt template and verbalizer are set as follows: 
\[
\begin{aligned}
    & \text{Sentence 1 is \{premise\},} \\
    & \text{sentence 2 is \{hypothesis\}.} \\
    & \text{They are [MASK].}
\end{aligned}
\]
\[
V = \left\{
\begin{aligned}
    & \text{entailment} \rightarrow \text{``entailed''},\; \\
    & \text{neutral} \rightarrow \text{``neutral''},\; \\
    & \text{contradiction} \rightarrow \text{``contradictory''}
\end{aligned}
\right\}
\]

\subsection{Hyperparameter Settings}

Hyperparameters are carefully tuned through empirical tests for optimal performance and computational efficiency. Based on preliminary experiments, we set the learning rate to \(1\times 10^{-5}\), batch size to 16 for all experiments. For the main experiment and few-shot task, each model is trained for 5 epochs. For detailed analysis we only train the model for 1 epoch. The optimizer used is AdamW~\citep{loshchilov2017decoupled} coupled with a linear scheduler (no warm-up steps). 

For partition sampling strategies (PME and PMD), we set the batch partitions in a 6:4 ratio (9 samples in the first partition and 7 samples in the second). 

Model selection for evaluation on the test set is based on the highest validation accuracy achieved during training.

During training, we maintain the same hyperparameters across all six sampling strategies and three experimental setups to ensure consistency in comparison. To mitigate the impact of random variation, we conduct each experiment using three different random seeds \{66, 88, 99\} and report the averaged results. For detailed analysis we use the result of seed 66.
All experiments are conducted using NVIDIA GeForce GTX 1080 Ti GPUs with 11 GB of memory. The entire pipeline is implemented using the PyTorch framework, which facilitated efficient training and evaluation.

\section{Reproducibility}\seclabel{repro}

The code for data processing and model training is available at the following Github repository: https://github.com/alitanokiki/self-adaptive-curriculum-nlu-acl2025.

\section{Detailed Analysis}\seclabel{result_details}

This section presents the results of all detailed analyses that were not included in the main text.

\begin{figure}[H]   
    \centering
    \includegraphics[width=\linewidth]{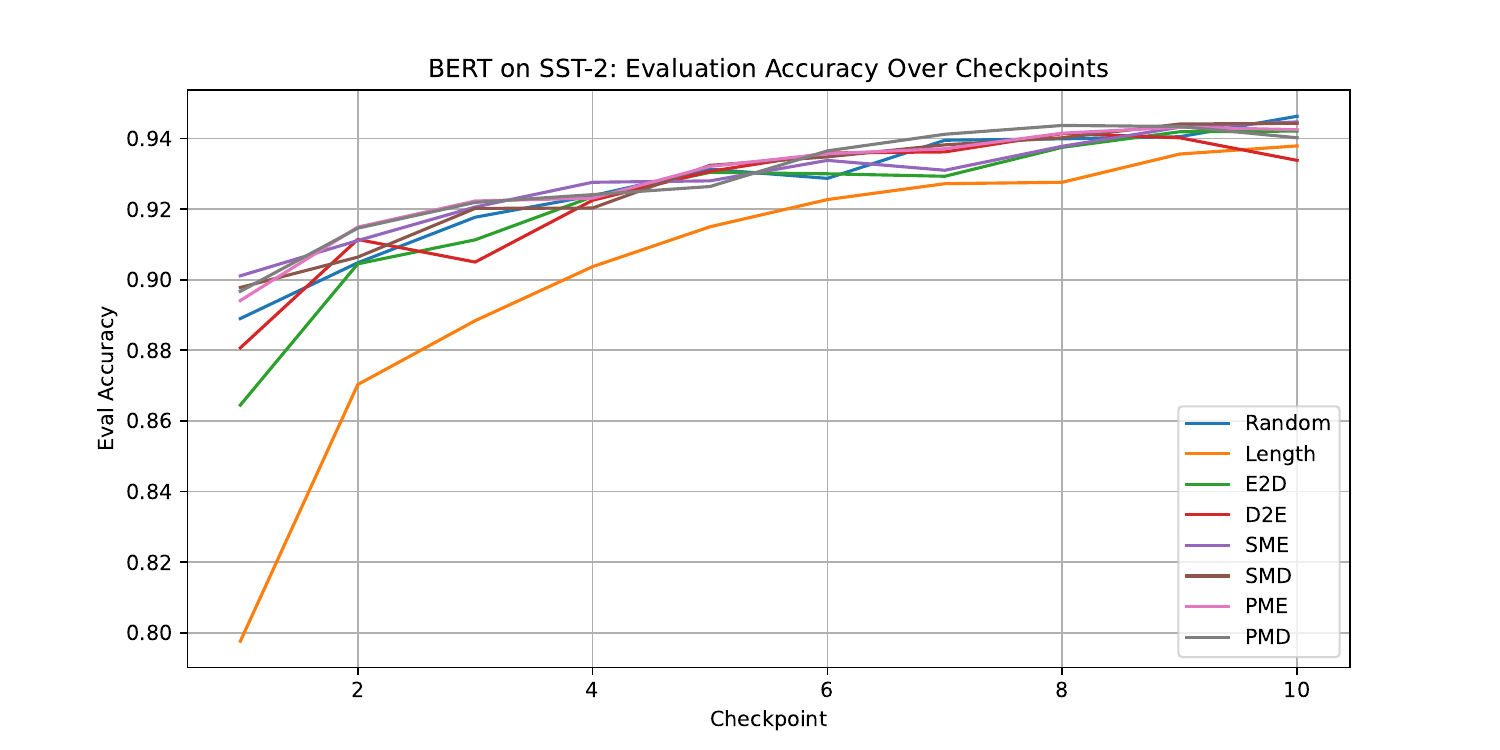}
    \caption{Average evaluation accuracy on BERT recorded at 10 checkpoints during a single epoch on SST-2.
}
    \label{fig:bertsst2acc}
\end{figure}

\begin{figure}[H]   
    \centering
    \includegraphics[width=\linewidth]{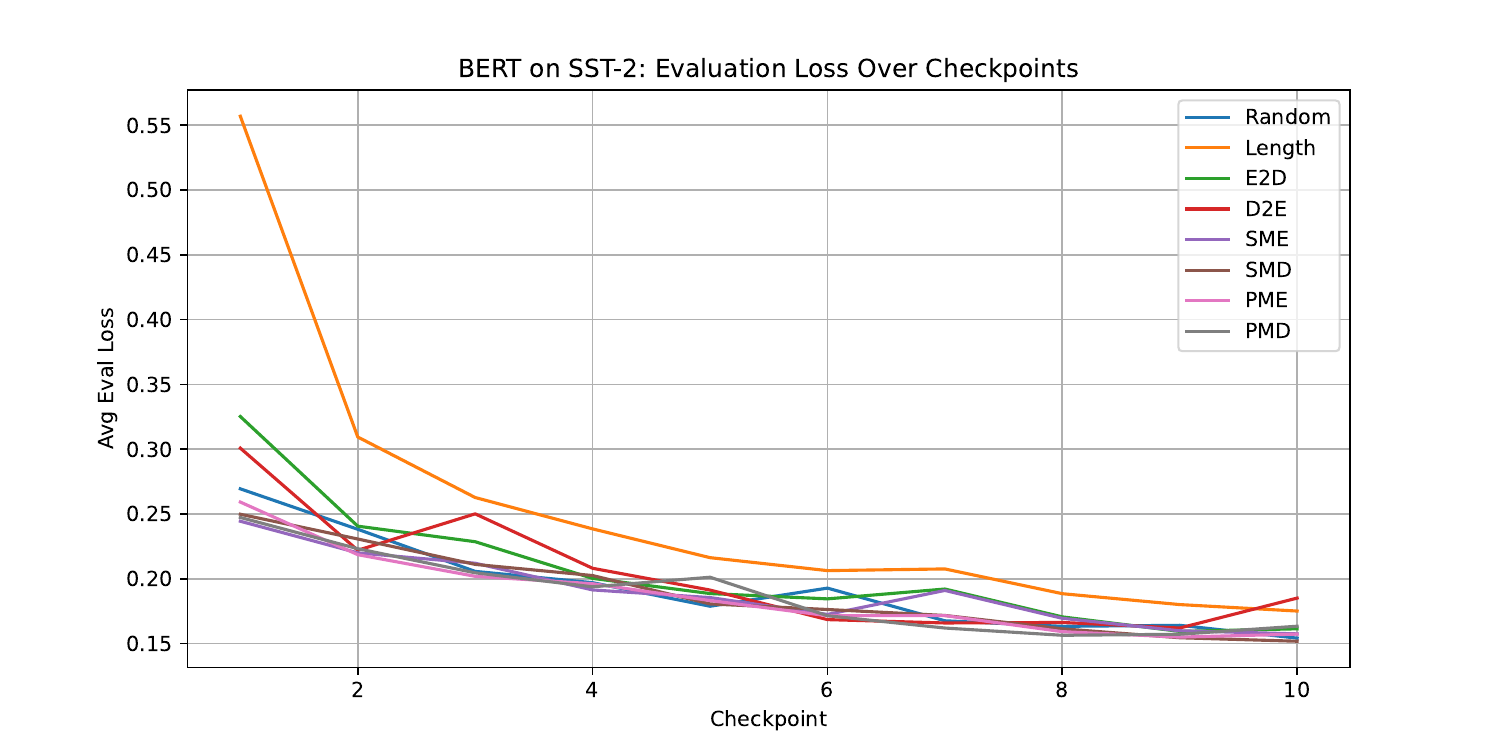}
    \caption{Average evaluation loss on BERT recorded at 10 checkpoints during a single epoch on SST-2.
}
    \label{fig:bertsst2loss}
\end{figure}

As shown in Figure \ref{fig:bertsst2acc} and \ref{fig:bertsst2loss}, probabilistic sampling methods (SME, SMD, PME, PMD) generally perform better.

\begin{figure}[H]   
    \centering
    \includegraphics[width=\linewidth]{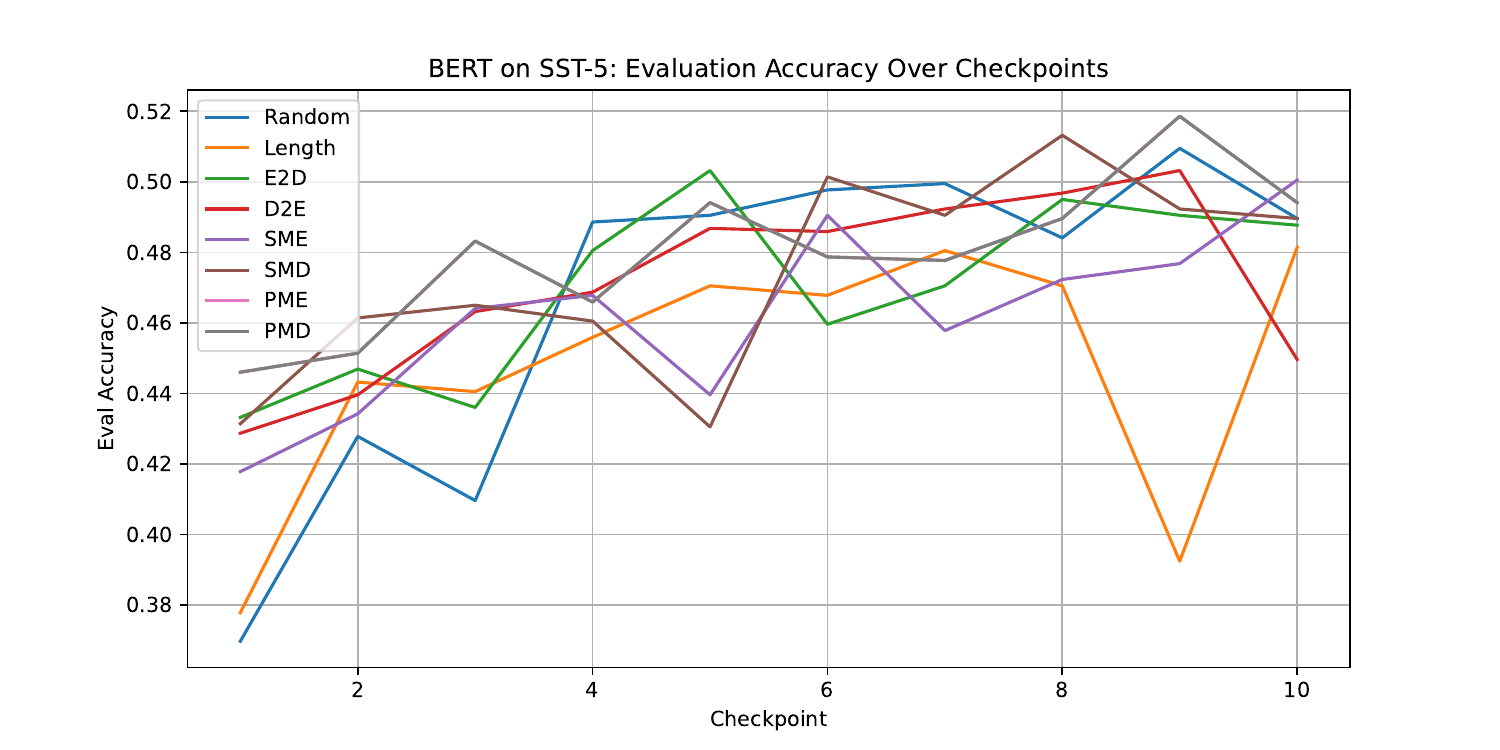}
    \caption{Average evaluation accuracy on BERT recorded at 10 checkpoints during a single epoch on SST-5. 
}
    \label{fig:bertsst5acc}
\end{figure}

\begin{figure}[H]   
    \centering
    \includegraphics[width=\linewidth]{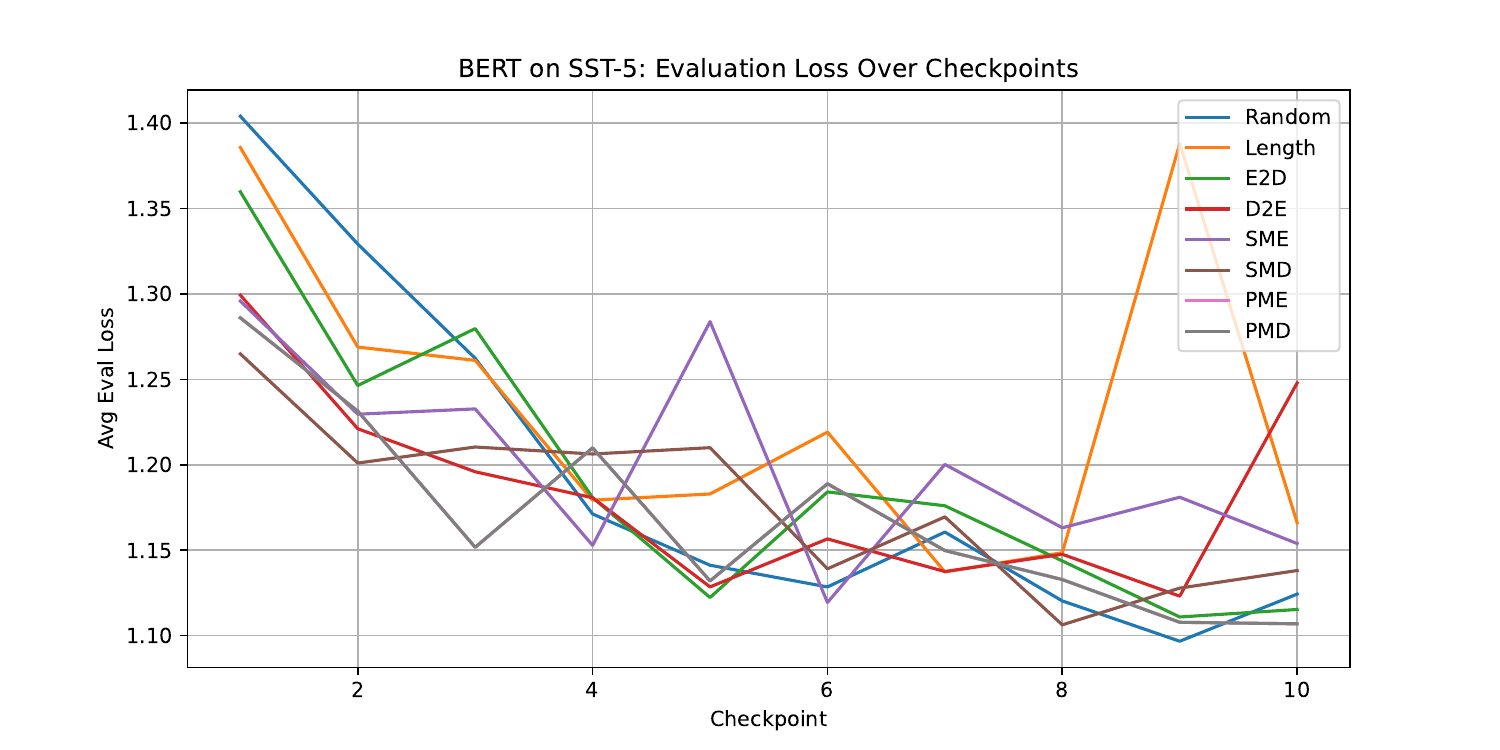}
    \caption{Average evaluation loss on BERT recorded at 10 checkpoints during a single epoch on SST-5.
}
    \label{fig:bertsst5loss}
\end{figure}

Figure \ref{fig:bertsst5acc} shows that all our training strategies start with strong performance. Performance fluctuates across strategies, with D2E performing significantly worse at the end. According to Figure \ref{fig:bertsst5loss}, SME achieves high accuracy but also results in higher loss.

\begin{figure}[H] 
    \centering
    \includegraphics[width=\linewidth]{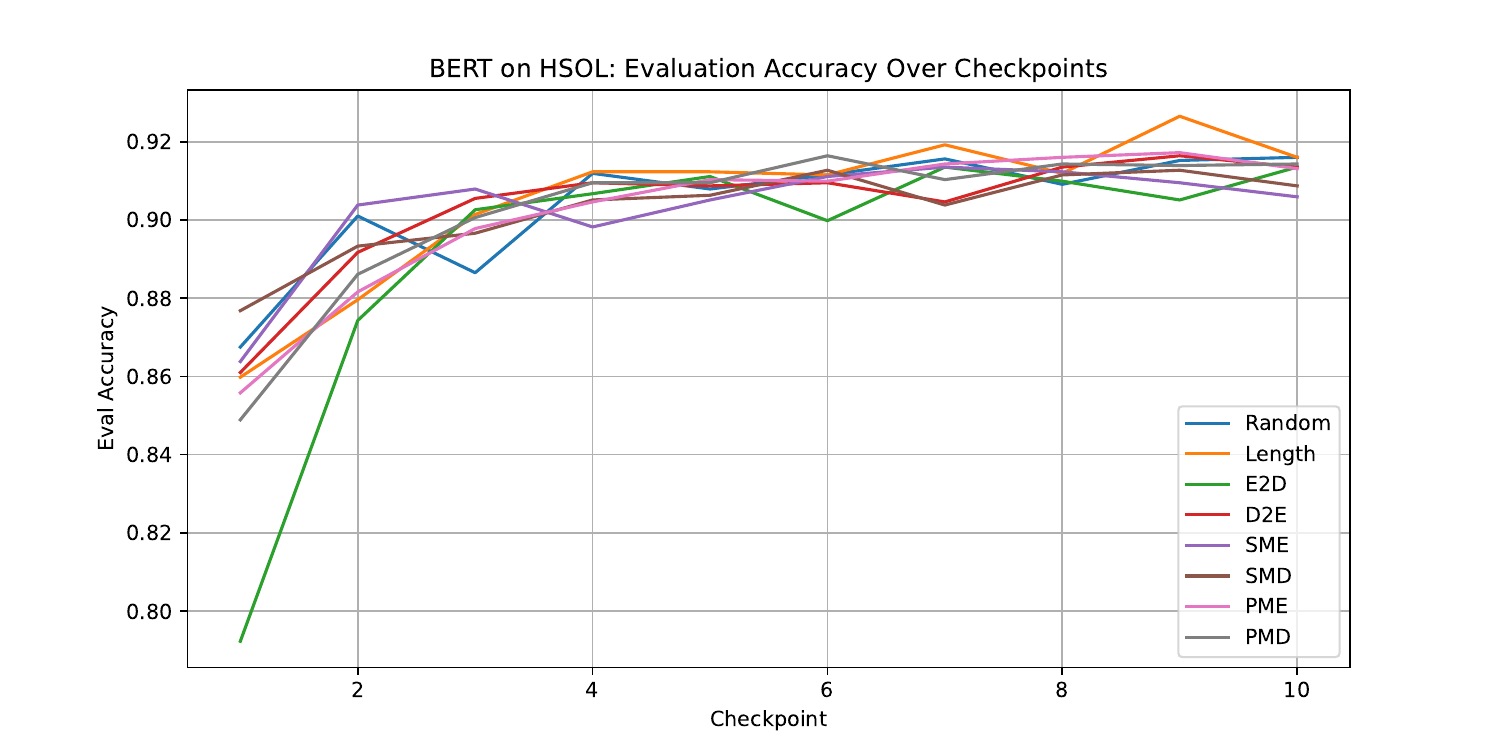}
    \caption{Average evaluation accuracy on BERT recorded at 10 checkpoints during a single epoch on HSOL. 
}
    \label{fig:berthsolacc}
\end{figure}

\begin{figure}[H]   
    \centering
    \includegraphics[width=\linewidth]{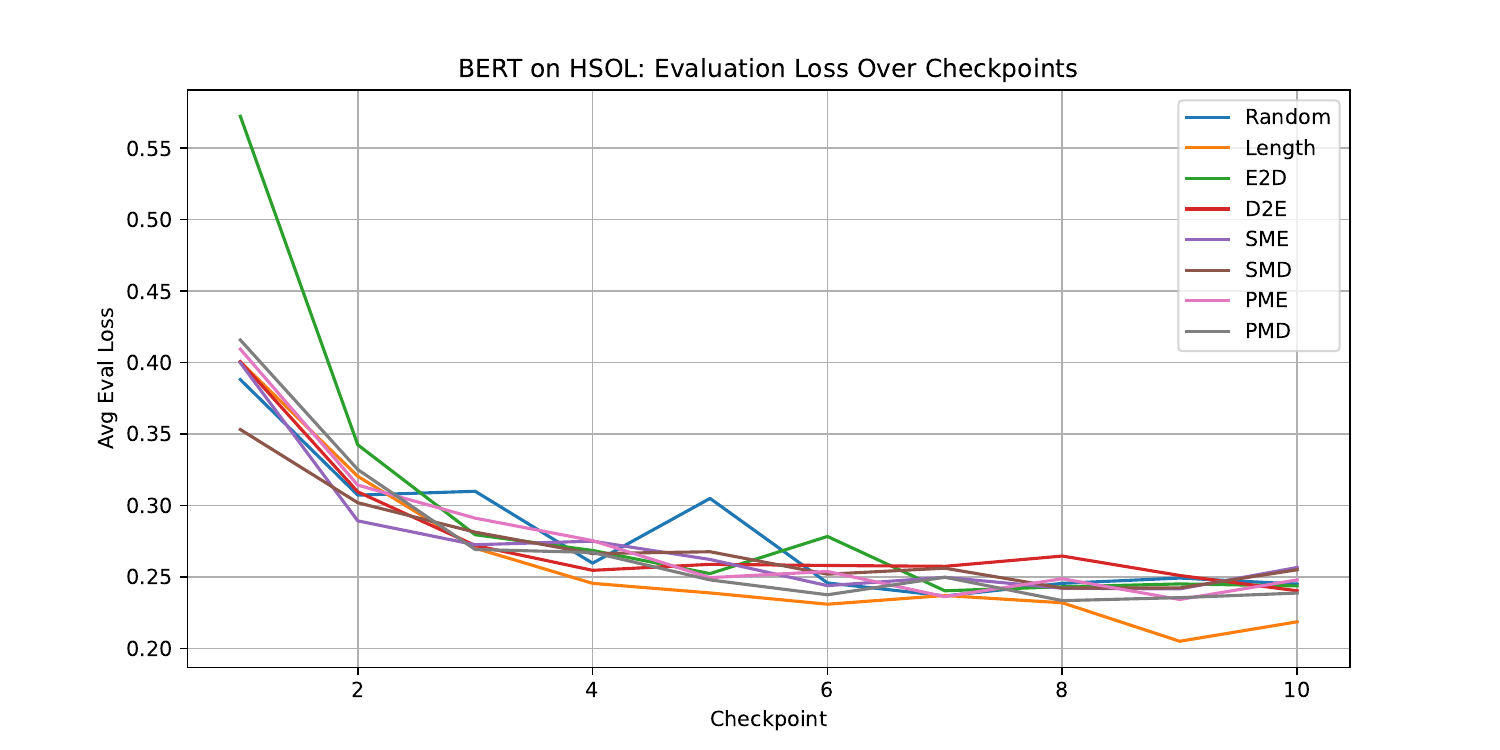}
    \caption{Average evaluation loss on BERT recorded at 10 checkpoints during a single epoch on HSOL.
}
    \label{fig:berthsolloss}
\end{figure}

Figure \ref{fig:berthsolacc} and \ref{fig:berthsolloss} indicate that E2D performs poorly at the beginning on imbalanced datasets. It is evident that after one epoch, our strategies no longer outperform the two baselines.

\begin{figure}[H]   
    \centering
    \includegraphics[width=\linewidth]{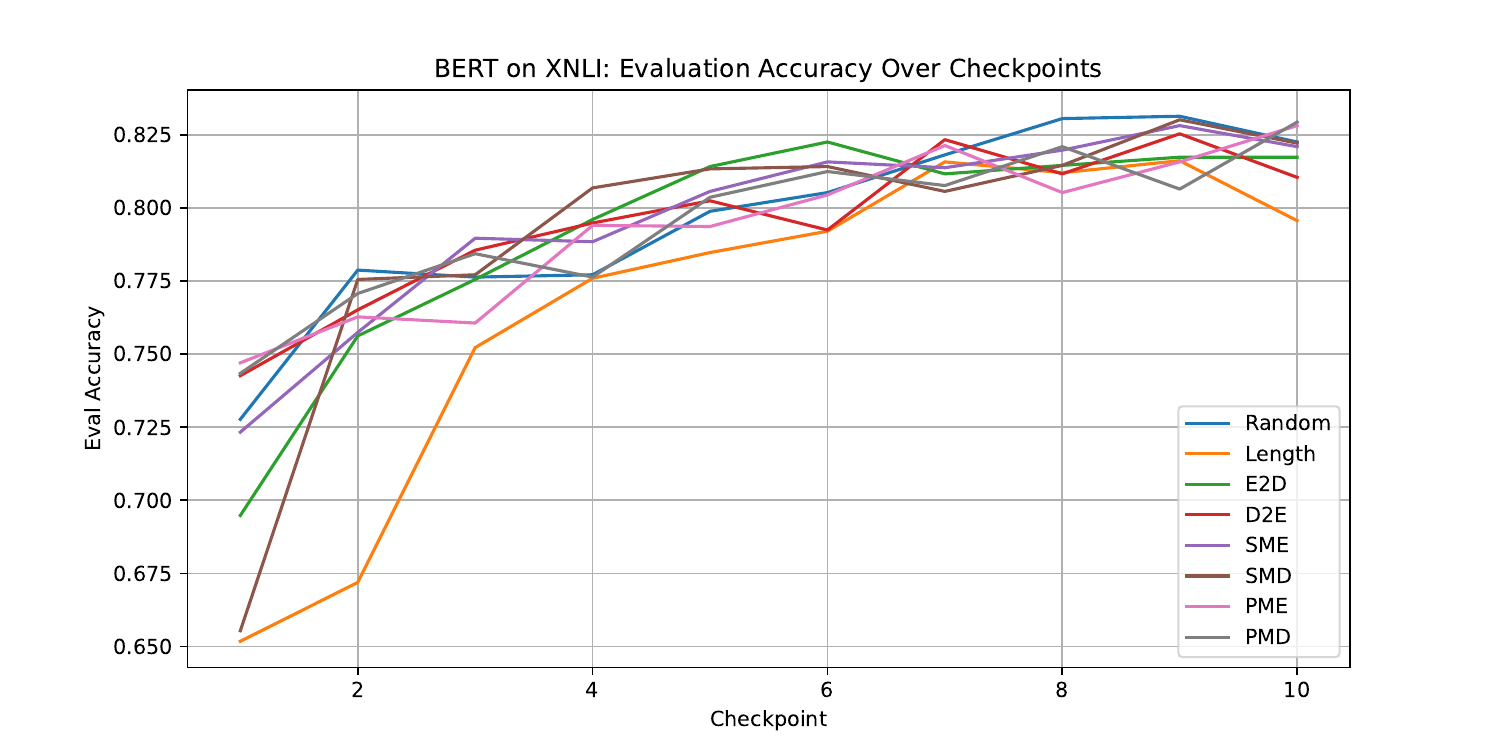}
    \caption{Average evaluation accuracy on BERT recorded at 10 checkpoints during a single epoch on XNLI. 
}
    \label{bertxnliacc}
\end{figure}

\begin{figure}[H]   
    \centering
    \includegraphics[width=\linewidth]{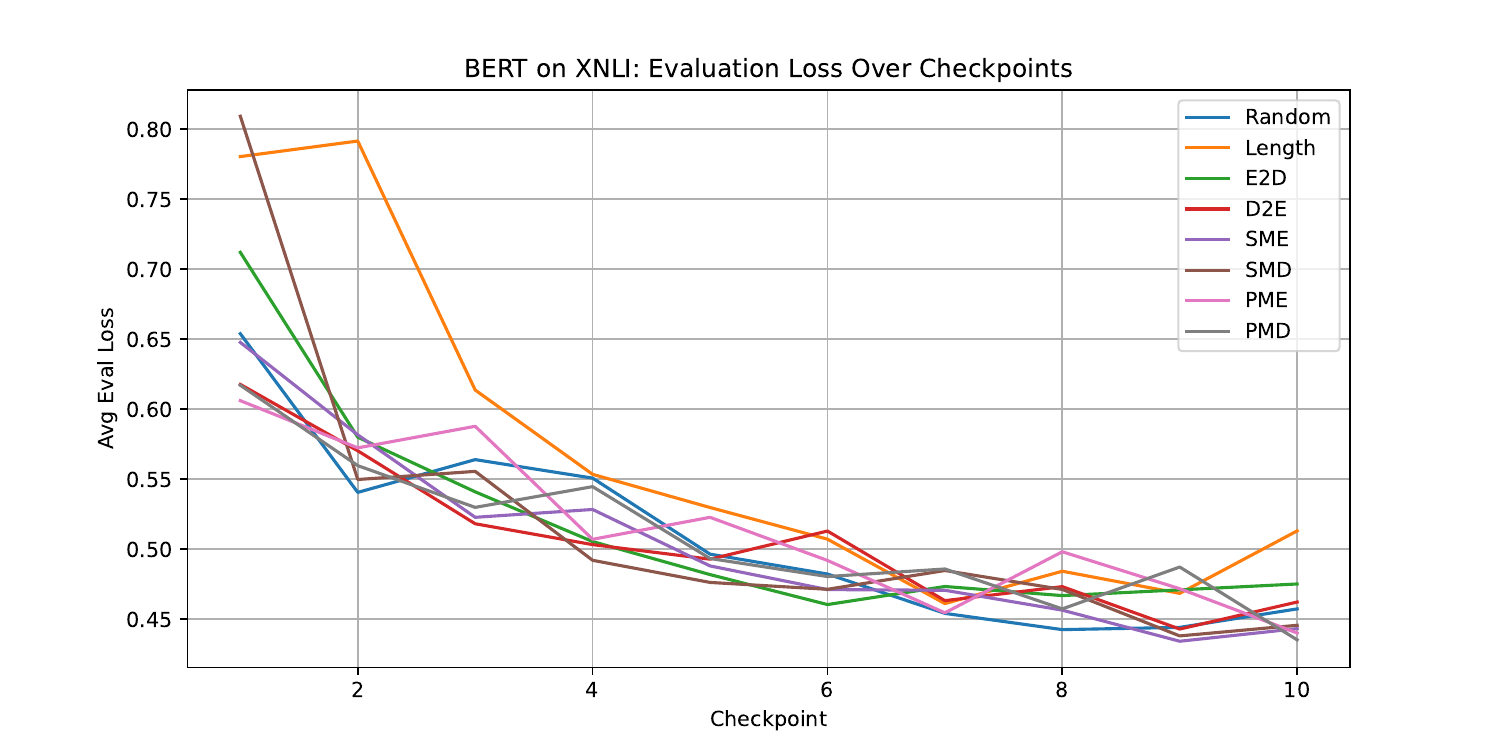}
    \caption{Average evaluation loss on BERT recorded at 10 checkpoints during a single epoch on XNLI.
}
    \label{fig:bertxnliloss}
\end{figure}

As shown in Figure \ref{bertxnliacc} and \ref{fig:bertxnliloss}, SMD starts off weaker but converges quickly. All probabilistic sampling methods (SME, SMD, PME, PMD) perform well in the end.

\begin{figure}[H]   
    \centering
    \includegraphics[width=\linewidth]{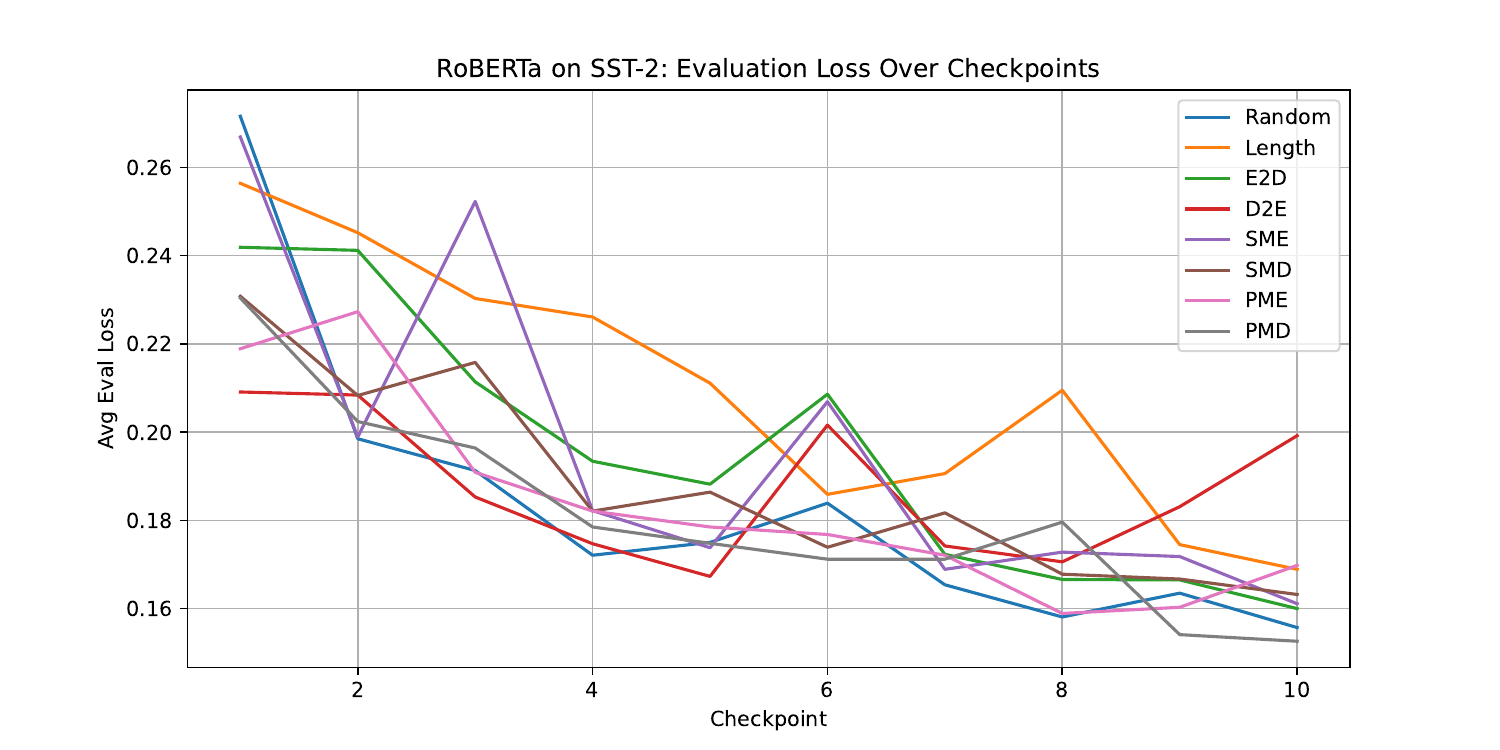}
    \caption{Average evaluation loss on RoBERTa recorded at 10 checkpoints during a single epoch on SST-2. 
}
    \label{fig:robertasst2loss}
\end{figure}

From Figure \ref{fig:robertasst2loss}, we see that D2E has low initial loss, but ends with the highest loss after one epoch.

\begin{figure}[H]   
    \centering
    \includegraphics[width=\linewidth]{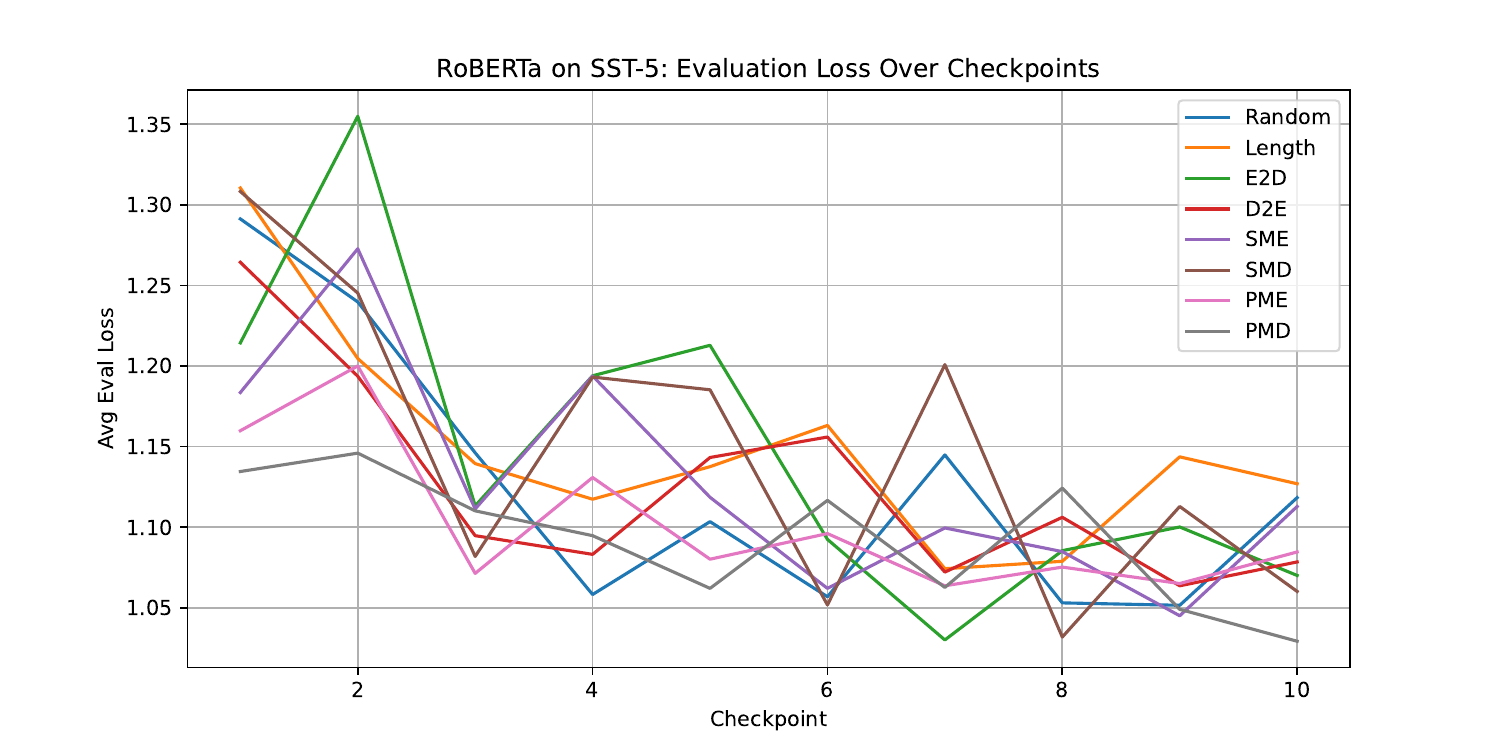}
    \caption{Average evaluation loss on RoBERTa recorded at 10 checkpoints during a single epoch on SST-5. 
}
    \label{fig:robertasst5loss}
\end{figure}

As shown in Figure \ref{fig:robertasst5loss}, PMD maintains the lowest and most stable loss throughout training.

\begin{figure}[H] 
    \centering
    \includegraphics[width=\linewidth]{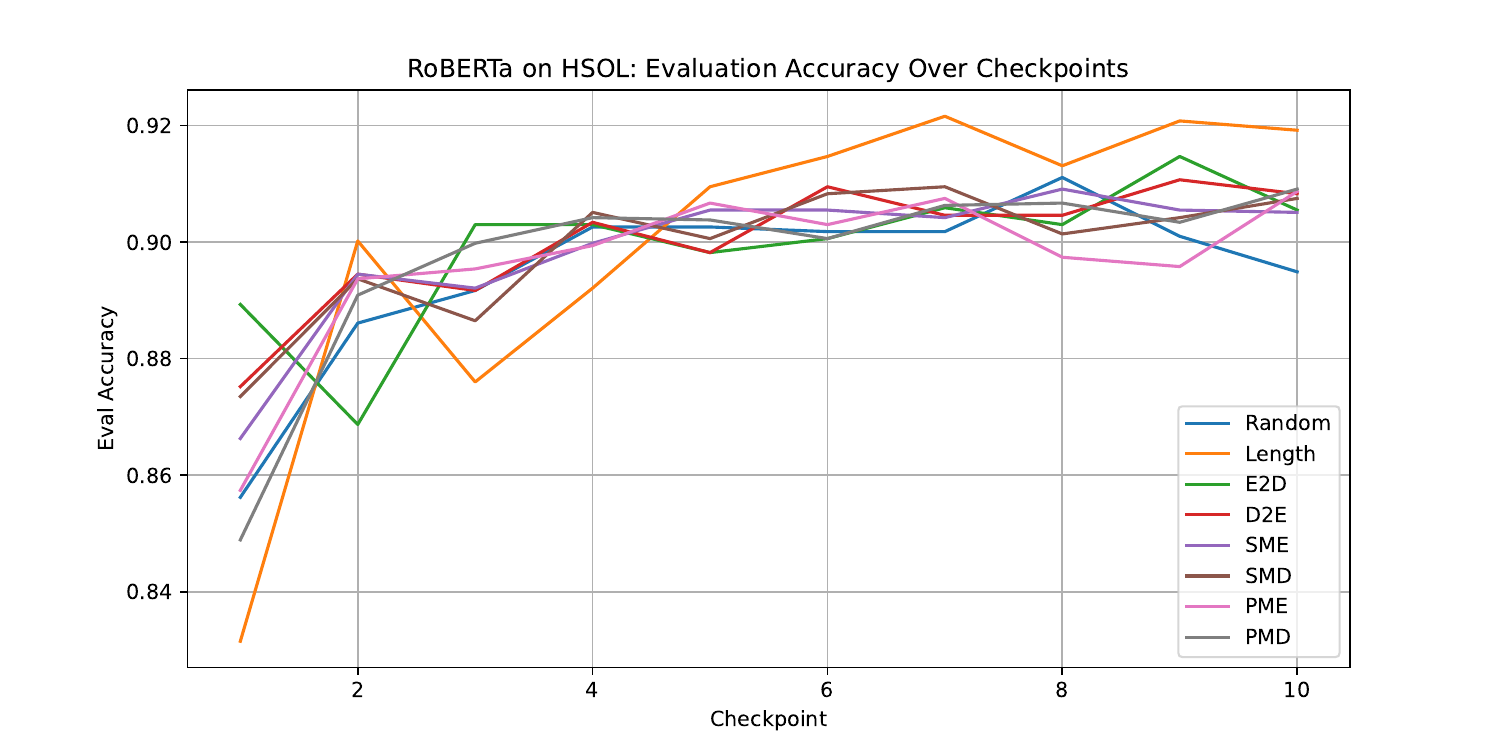}
    \caption{Average evaluation accuracy on RoBERTa recorded at 10 checkpoints during a single epoch on HSOL. 
}
    \label{fig:robertahsolacc}
\end{figure}

Figure \ref{fig:robertahsolacc} reveals that E2D shows early advantages, but the Length baseline performs best in the final stage.

\begin{figure}[H]   
    \centering
    \includegraphics[width=\linewidth]{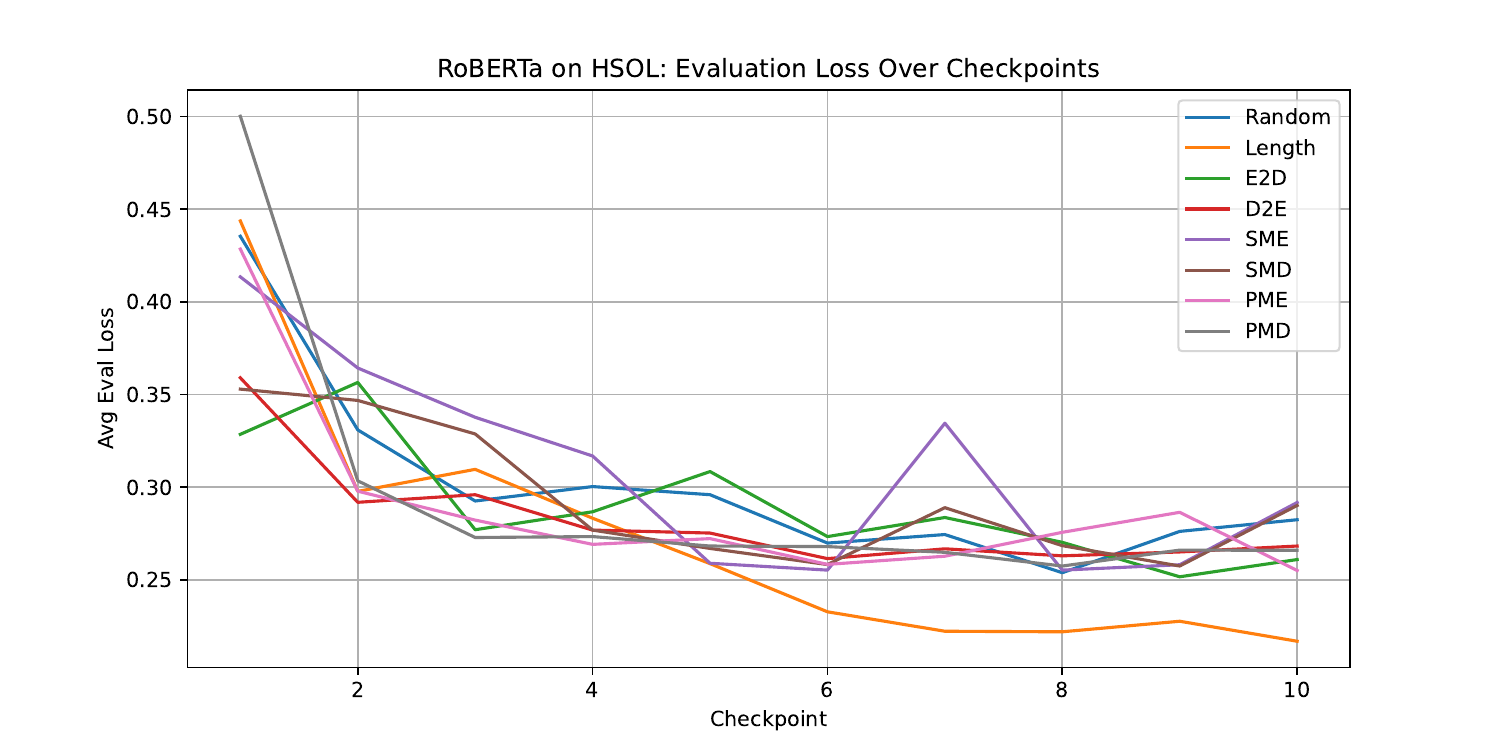}
    \caption{Average evaluation loss on RoBERTa recorded at 10 checkpoints during a single epoch on HSOL. 
}
    \label{fig:robertahsolloss}
\end{figure}

According to Figure \ref{fig:robertahsolloss}, PMD initially has the highest loss, but it decreases rapidly.

\begin{figure}[H]   
    \centering
    \includegraphics[width=\linewidth]{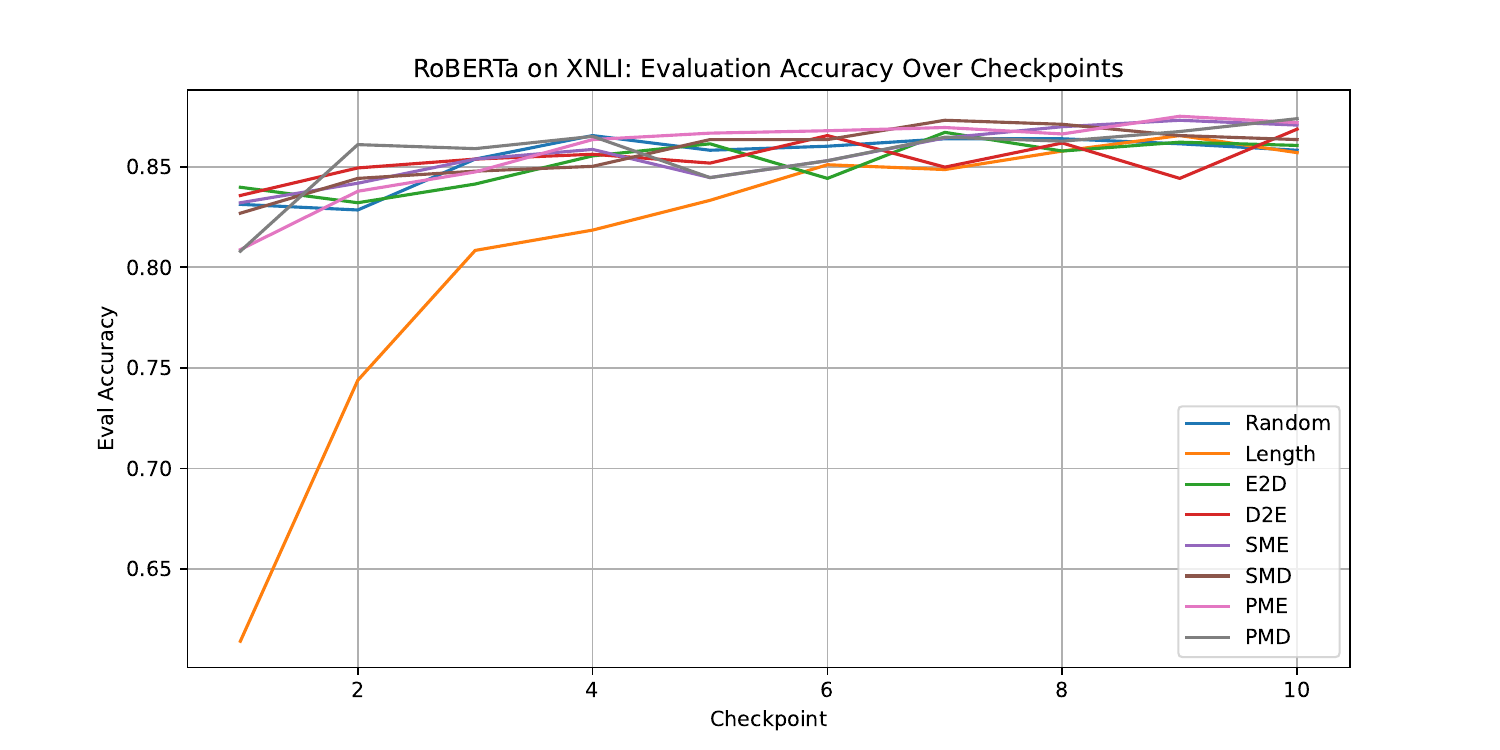}
    \caption{Average evaluation accuracy on RoBERTa recorded at 10 checkpoints during a single epoch on XNLI. }
    \label{robertaxnliacc}
\end{figure}

\begin{figure}[H]   
    \centering
    \includegraphics[width=\linewidth]{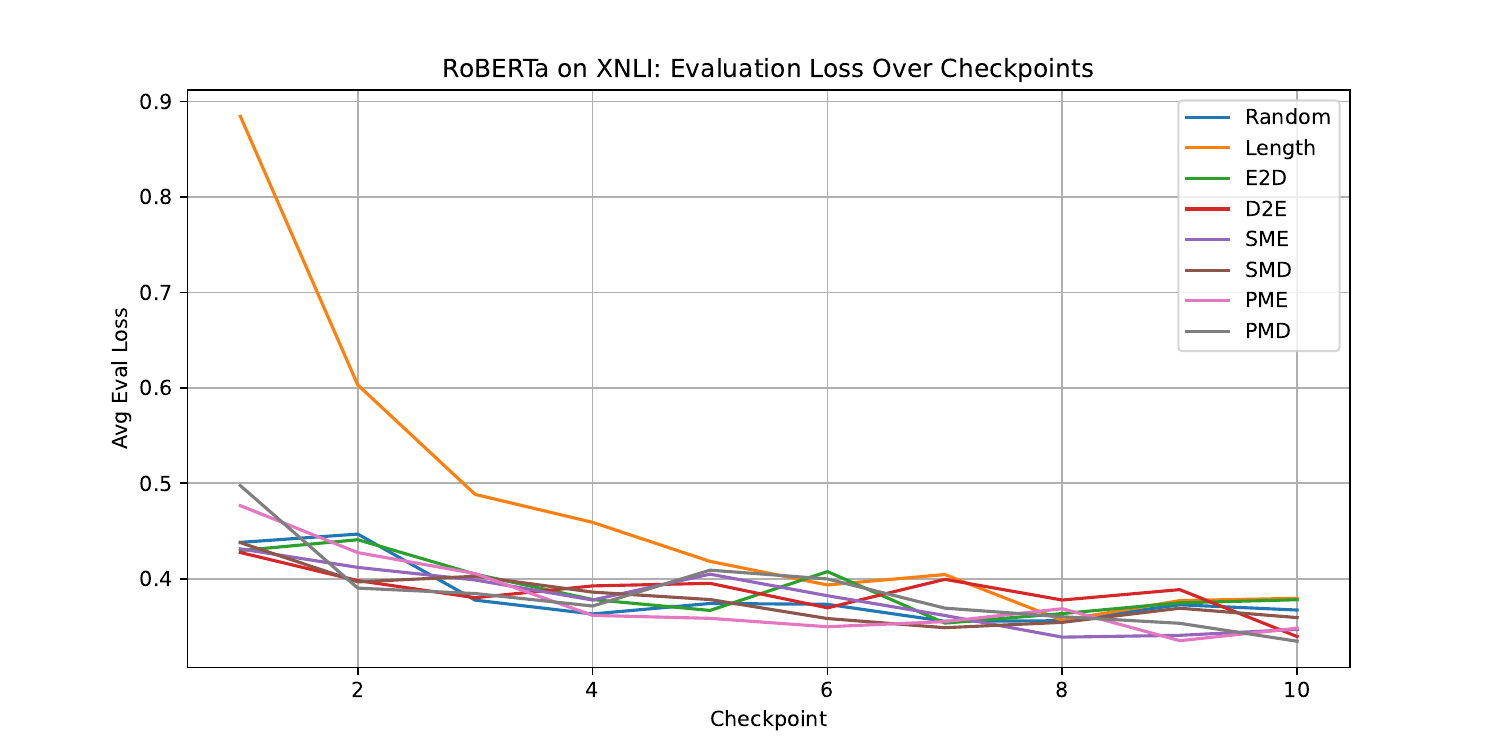}
    \caption{Average evaluation loss on RoBERTa recorded at 10 checkpoints during a single epoch on XNLI. 
}
    \label{fig:robertaxnliloss}
\end{figure}

Figure \ref{robertaxnliacc} and \ref{fig:robertaxnliloss} show that apart from the baseline Length, differences in performance across methods are minor.

\section{Difficulty Score Distribution Over Training Time}\seclabel{score_distribution}

We analyze the evolution of sample difficulty score distributions under various training strategies across different datasets, using both BERT and RoBERTa models. While different strategies exhibit similar trends within the same dataset, the distributional patterns vary notably across datasets. Due to the consistency observed within each dataset, we take the BERT model as a representative example to illustrate these trends. Specifically, we present the score distribution changes of BERT trained with the baseline Random on each dataset, highlighting how dataset characteristics influence learning dynamics.

\begin{figure}[H]   
    \centering
    \includegraphics[width=\linewidth]{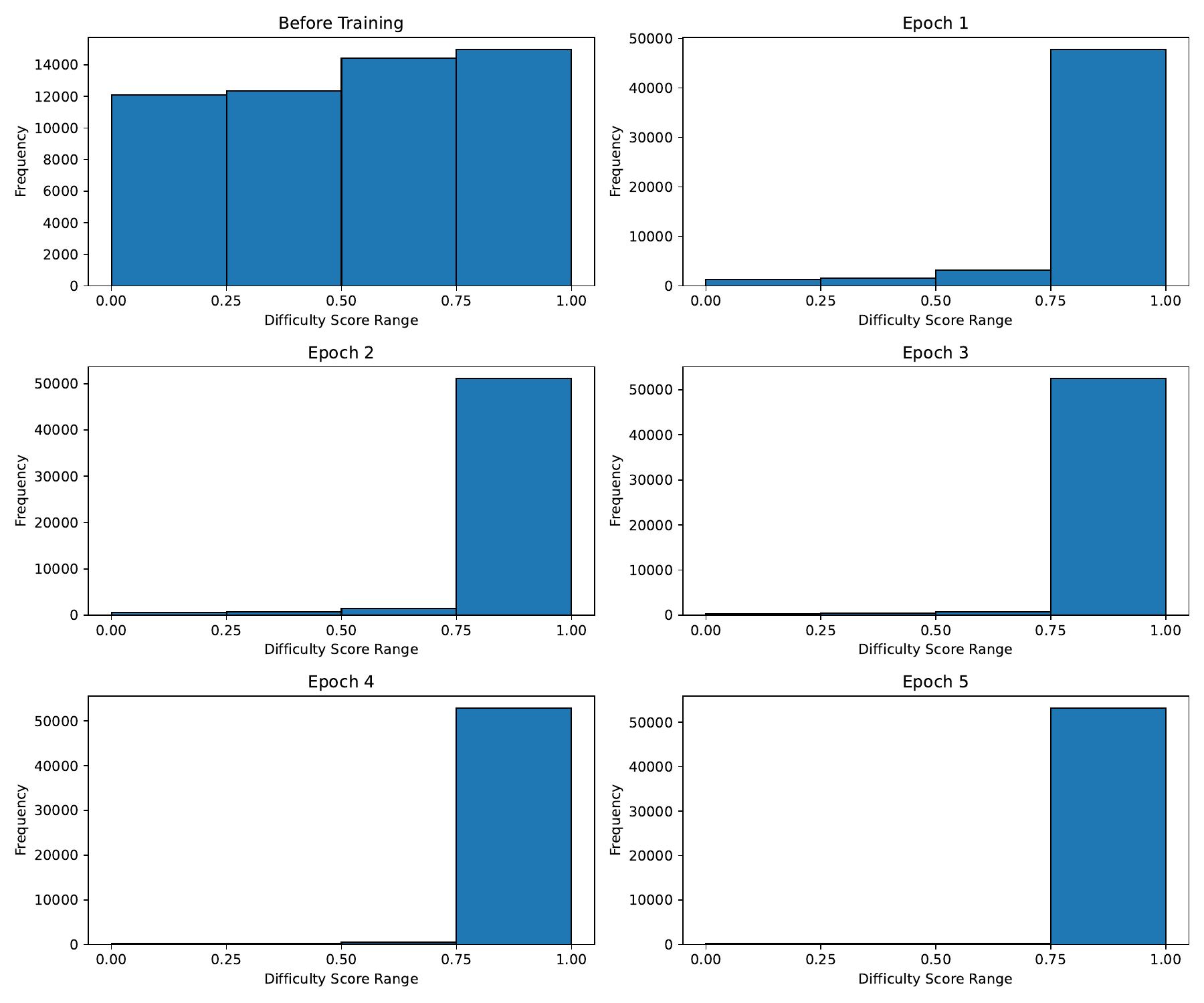}
    \caption{Sample difficulty score distributions on SST-2 before training and after each of five training epochs using BERT.
}
    \label{fig: distribution_sst2}
\end{figure}

As shown in Figure \ref{fig: distribution_sst2}, the initial difficulty score distribution on the SST-2 dataset is relatively uniform. After the first epoch, the number of easy samples increases sharply, indicating that the model has learned substantially during the initial phase. The shift toward higher scores suggests increased model confidence. In subsequent epochs, the distribution stabilizes, reflecting more consistent learning dynamics.

\begin{figure}[H]   
    \centering
    \includegraphics[width=\linewidth]{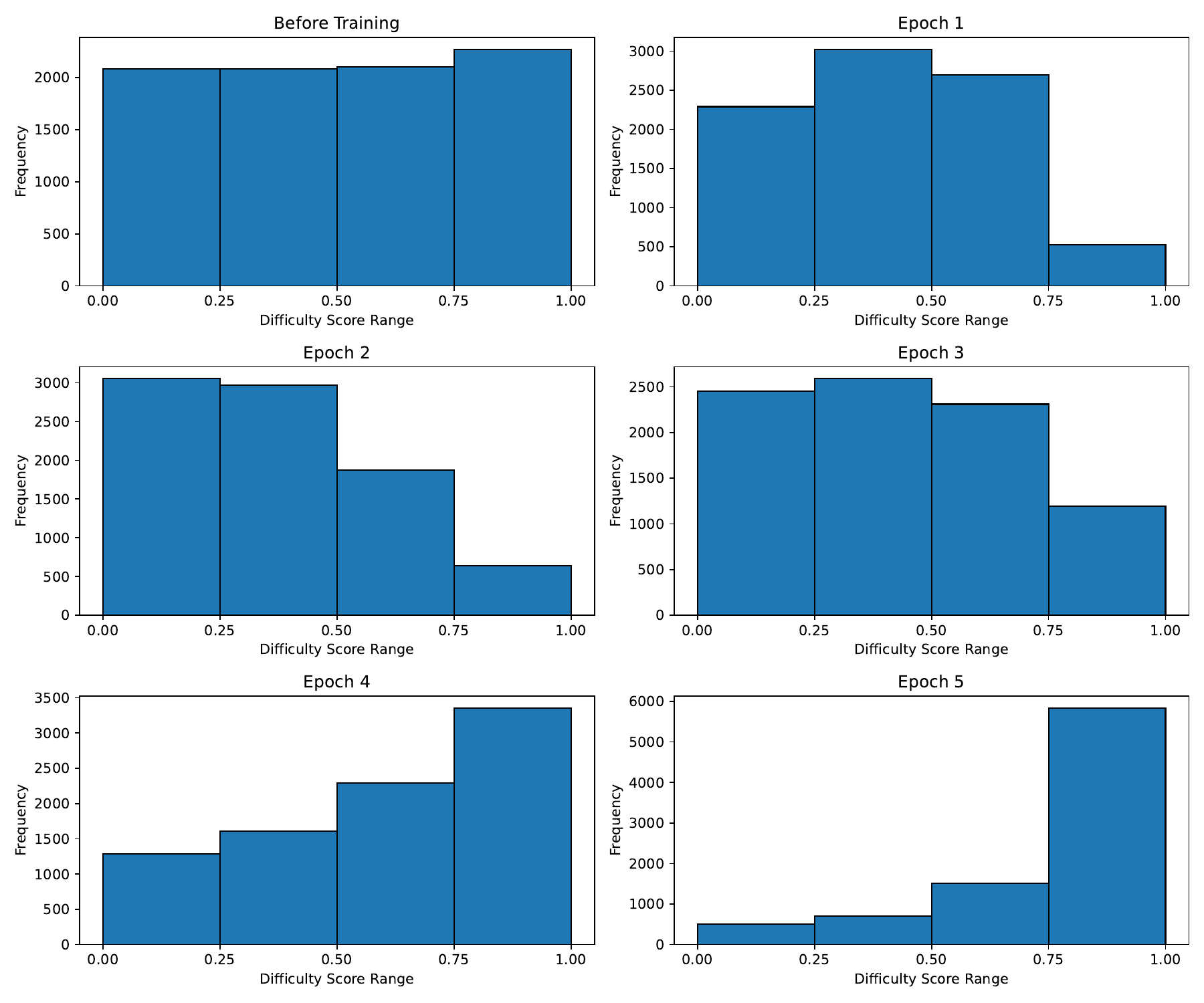}
    \caption{Sample difficulty score distributions on SST-5 before training and after each of five training epochs using BERT.
}
    \label{fig: distribution_sst5}
\end{figure}

Figure \ref{fig: distribution_sst5} shows the evolution of difficulty score distribution for the BERT model on the SST-5 dataset. After one epoch, the number of relatively difficult samples increases, which may be attributed to the way difficulty scores are computed. One possible explanation is that, for multi-class classification, the difficulty score is defined as the absolute difference between the top two class probabilities. In this dataset, certain samples may have high but very close probabilities for adjacent sentiment classes, such as “negative” and “very negative” or “positive” and “very positive.” As the model begins to learn useful features, the score difference of these low-confidence difficult samples tends to increase. Once the model has acquired more discriminative features, it becomes easier to correctly classify these borderline cases, resulting in higher overall accuracy. In this sense, low-confidence difficult samples may be the easiest to convert from incorrect to correct predictions. This interpretation is further supported by the observed score distribution, indicating that the model learned meaningful features within the first epoch.

\begin{figure}[H]   
    \centering
    \includegraphics[width=\linewidth]{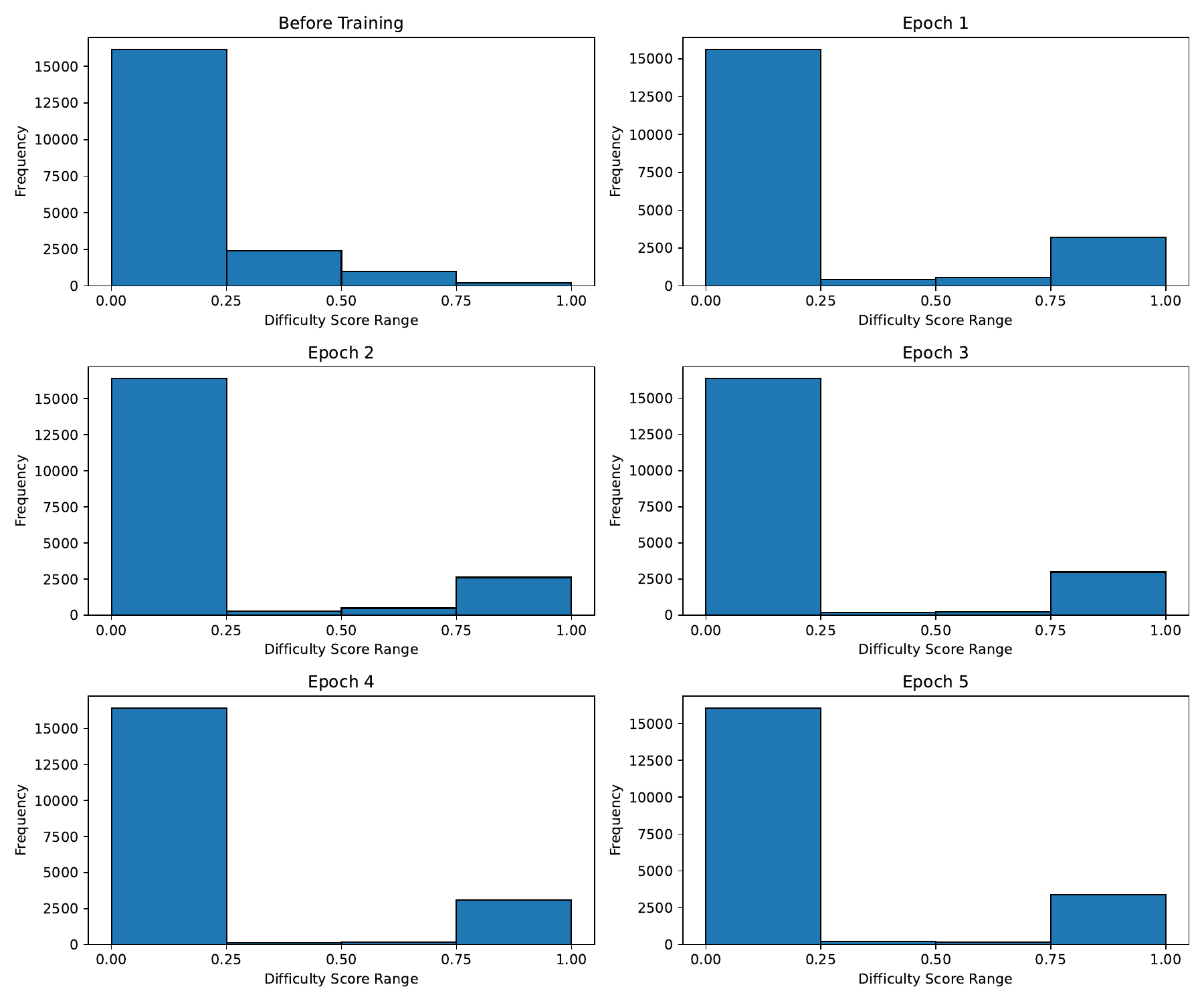}
    \caption{Sample difficulty score distributions on HSOL before training and after each of five training epochs using BERT.
}
    \label{fig: distribution_hso}
\end{figure}

As shown in Figure \ref{fig: distribution_hso}, the HSOL dataset is highly imbalanced both in terms of label distribution and initial difficulty scores, with a large proportion of hard samples. After one training epoch, the number of easy samples increases slightly, indicating some initial learning progress. However, even after training is completed, a substantial number of difficult samples remain, suggesting that the model struggles to learn from a significant portion of the data.

\begin{figure}[H]   
    \centering
    \includegraphics[width=\linewidth]{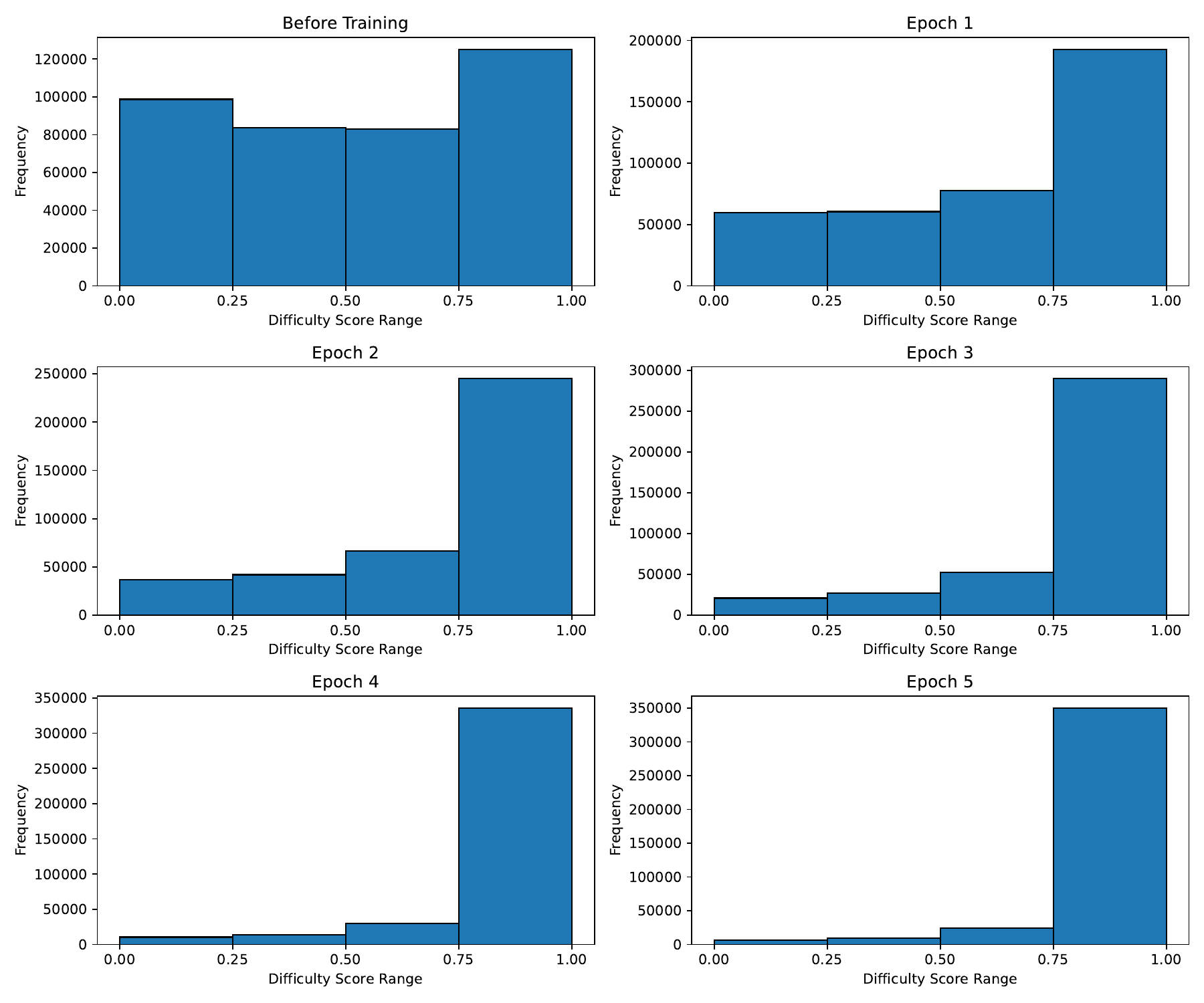}
    \caption{Sample difficulty score distributions on XNLI before training and after each of five training epochs using BERT.
}
    \label{fig: distribution_xnli}
\end{figure}

As shown in Figure \ref{fig: distribution_xnli}, the XNLI dataset exhibits a relatively balanced initial distribution of difficulty scores. Throughout training, both easy and difficult samples gradually increase or decrease in number in a stable manner, indicating consistent learning dynamics. This stable progression may be attributed to the large size and diversity of the dataset, which provides sufficient training signals across difficulty levels.

\end{document}